%% file: main.tex
\documentclass[10pt,twocolumn,letterpaper]{article}

\usepackage[pagenumbers]{cvpr}      


\usepackage{multirow}
\usepackage{enumitem}
\usepackage{float}
\usepackage{adjustbox}
\usepackage{placeins}
\usepackage{needspace}
\usepackage{tikz}
\usetikzlibrary{arrows.meta,decorations.pathreplacing,fit,backgrounds,calc,positioning}
\definecolor{vdmblue}{HTML}{2A78D6}
\definecolor{vdmorange}{HTML}{EB6834}
\definecolor{vdmaqua}{HTML}{1BAF7A}

\definecolor{cvprblue}{rgb}{0.21,0.49,0.74}
\usepackage[pagebackref,breaklinks,colorlinks,allcolors=cvprblue,hypertexnames=false]{hyperref}

\def\confName{CVPR}
\def\confYear{2027}

\input{generated/macros}

\input{generated/v3/macros_v3}

\title{Visual Jev: Accurate and Efficient Decisions\\
  from Shared Visual Context}

\usetikzlibrary{svg.path}
\newcommand{\projecticon}[1]{\raisebox{-0.16em}{%
  \begin{tikzpicture}[x=0.35pt,y=-0.35pt]
  \path[use as bounding box] (0,0) rectangle (24,24);
  #1
  \end{tikzpicture}}}
\newcommand{\websiteicon}{\projecticon{%
  \draw[cvprblue,line width=0.55pt] (12,12) circle (10);
  \draw[cvprblue,line width=0.45pt] (12,12) ellipse (4.5 and 10);
  \draw[cvprblue,line width=0.45pt] (2,12)--(22,12) (4,6.5)--(20,6.5) (4,17.5)--(20,17.5);}}
\newcommand{\githubicon}{\projecticon{\begin{scope}[scale=0.35,yscale=-1]\path[fill=black] svg {M12 .297c-6.63 0-12 5.373-12 12 0 5.303 3.438 9.8 8.205 11.385.6.113.82-.258.82-.577 0-.285-.01-1.04-.015-2.04-3.338.724-4.042-1.61-4.042-1.61C4.422 18.07 3.633 17.7 3.633 17.7c-1.087-.744.084-.729.084-.729 1.205.084 1.838 1.236 1.838 1.236 1.07 1.835 2.809 1.305 3.495.998.108-.776.417-1.305.76-1.605-2.665-.3-5.466-1.332-5.466-5.93 0-1.31.465-2.38 1.235-3.22-.135-.303-.54-1.523.105-3.176 0 0 1.005-.322 3.3 1.23.96-.267 1.98-.399 3-.405 1.02.006 2.04.138 3 .405 2.28-1.552 3.285-1.23 3.285-1.23.645 1.653.24 2.873.12 3.176.765.84 1.23 1.91 1.23 3.22 0 4.61-2.805 5.625-5.475 5.92.42.36.81 1.096.81 2.22 0 1.606-.015 2.896-.015 3.286 0 .315.21.69.825.57C20.565 22.092 24 17.592 24 12.297c0-6.627-5.373-12-12-12};\end{scope}}}
\definecolor{hfcolorffffff}{HTML}{ffffff}
\definecolor{hfcolorFFD21E}{HTML}{FFD21E}
\definecolor{hfcolorFF9D0B}{HTML}{FF9D0B}
\definecolor{hfcolor3A3B45}{HTML}{3A3B45}
\definecolor{hfcolorFF323D}{HTML}{FF323D}
\newcommand{\huggingfaceicon}{\projecticon{\begin{scope}[scale=0.08842,yscale=-1]\path[fill=hfcolorffffff] svg {M94.25 70.08a8.28 8.28 0 0 1-.43 6.46 10.57 10.57 0 0 1-3 3.6 25.18 25.18 0 0 1-5.7 3.2 65.74 65.74 0 0 1-7.56 2.65 46.67 46.67 0 0 1-11.42 1.68c-5.42.05-10.09-1.23-13.4-4.5a40.4 40.4 0 0 1-10.14.03c-3.34 3.25-7.99 4.52-13.39 4.47a46.82 46.82 0 0 1-11.43-1.68 66.37 66.37 0 0 1-7.55-2.65c-2.28-.98-4.17-2-5.68-3.2a10.5 10.5 0 0 1-3.02-3.6c-.99-2-1.18-4.3-.42-6.46a8.54 8.54 0 0 1-.33-5.63c.25-.95.66-1.83 1.18-2.61a8.67 8.67 0 0 1 2.1-8.47 8.23 8.23 0 0 1 2.82-2.07 41.75 41.75 0 1 1 81.3-.12 8.27 8.27 0 0 1 3.11 2.19 8.7 8.7 0 0 1 2.1 8.47c.52.78.93 1.66 1.18 2.61a8.61 8.61 0 0 1-.32 5.63Z};\path[fill=hfcolorFFD21E] svg {M47.21 76.5a34.75 34.75 0 1 0 0-69.5 34.75 34.75 0 0 0 0 69.5Z};\path[fill=hfcolorFF9D0B] svg {M81.96 41.75a34.75 34.75 0 1 0-69.5 0 34.75 34.75 0 0 0 69.5 0Zm-73.5 0a38.75 38.75 0 1 1 77.5 0 38.75 38.75 0 0 1-77.5 0Z};\path[fill=hfcolor3A3B45] svg {M58.5 32.3c1.28.44 1.78 3.06 3.07 2.38a5 5 0 1 0-6.76-2.07c.61 1.15 2.55-.72 3.7-.32ZM34.95 32.3c-1.28.44-1.79 3.06-3.07 2.38a5 5 0 1 1 6.76-2.07c-.61 1.15-2.56-.72-3.7-.32Z};\path[fill=hfcolorFF323D] svg {M46.96 56.29c9.83 0 13-8.76 13-13.26 0-2.34-1.57-1.6-4.09-.36-2.33 1.15-5.46 2.74-8.9 2.74-7.19 0-13-6.88-13-2.38s3.16 13.26 13 13.26Z};\path[fill=hfcolor3A3B45,even odd rule] svg {M39.43 54a8.7 8.7 0 0 1 5.3-4.49c.4-.12.81.57 1.24 1.28.4.68.82 1.37 1.24 1.37.45 0 .9-.68 1.33-1.35.45-.7.89-1.38 1.32-1.25a8.61 8.61 0 0 1 5 4.17c3.73-2.94 5.1-7.74 5.1-10.7 0-2.34-1.57-1.6-4.09-.36l-.14.07c-2.31 1.15-5.39 2.67-8.77 2.67s-6.45-1.52-8.77-2.67c-2.6-1.29-4.23-2.1-4.23.29 0 3.05 1.46 8.06 5.47 10.97Z};\path[fill=hfcolorFF9D0B] svg {M70.71 37a3.25 3.25 0 1 0 0-6.5 3.25 3.25 0 0 0 0 6.5ZM24.21 37a3.25 3.25 0 1 0 0-6.5 3.25 3.25 0 0 0 0 6.5ZM17.52 48c-1.62 0-3.06.66-4.07 1.87a5.97 5.97 0 0 0-1.33 3.76 7.1 7.1 0 0 0-1.94-.3c-1.55 0-2.95.59-3.94 1.66a5.8 5.8 0 0 0-.8 7 5.3 5.3 0 0 0-1.79 2.82c-.24.9-.48 2.8.8 4.74a5.22 5.22 0 0 0-.37 5.02c1.02 2.32 3.57 4.14 8.52 6.1 3.07 1.22 5.89 2 5.91 2.01a44.33 44.33 0 0 0 10.93 1.6c5.86 0 10.05-1.8 12.46-5.34 3.88-5.69 3.33-10.9-1.7-15.92-2.77-2.78-4.62-6.87-5-7.77-.78-2.66-2.84-5.62-6.25-5.62a5.7 5.7 0 0 0-4.6 2.46c-1-1.26-1.98-2.25-2.86-2.82A7.4 7.4 0 0 0 17.52 48Zm0 4c.51 0 1.14.22 1.82.65 2.14 1.36 6.25 8.43 7.76 11.18.5.92 1.37 1.31 2.14 1.31 1.55 0 2.75-1.53.15-3.48-3.92-2.93-2.55-7.72-.68-8.01.08-.02.17-.02.24-.02 1.7 0 2.45 2.93 2.45 2.93s2.2 5.52 5.98 9.3c3.77 3.77 3.97 6.8 1.22 10.83-1.88 2.75-5.47 3.58-9.16 3.58-3.81 0-7.73-.9-9.92-1.46-.11-.03-13.45-3.8-11.76-7 .28-.54.75-.76 1.34-.76 2.38 0 6.7 3.54 8.57 3.54.41 0 .7-.17.83-.6.79-2.85-12.06-4.05-10.98-8.17.2-.73.71-1.02 1.44-1.02 3.14 0 10.2 5.53 11.68 5.53.11 0 .2-.03.24-.1.74-1.2.33-2.04-4.9-5.2-5.21-3.16-8.88-5.06-6.8-7.33.24-.26.58-.38 1-.38 3.17 0 10.66 6.82 10.66 6.82s2.02 2.1 3.25 2.1c.28 0 .52-.1.68-.38.86-1.46-8.06-8.22-8.56-11.01-.34-1.9.24-2.85 1.31-2.85Z};\path[fill=hfcolorFFD21E] svg {M38.6 76.69c2.75-4.04 2.55-7.07-1.22-10.84-3.78-3.77-5.98-9.3-5.98-9.3s-.82-3.2-2.69-2.9c-1.87.3-3.24 5.08.68 8.01 3.91 2.93-.78 4.92-2.29 2.17-1.5-2.75-5.62-9.82-7.76-11.18-2.13-1.35-3.63-.6-3.13 2.2.5 2.79 9.43 9.55 8.56 11-.87 1.47-3.93-1.71-3.93-1.71s-9.57-8.71-11.66-6.44c-2.08 2.27 1.59 4.17 6.8 7.33 5.23 3.16 5.64 4 4.9 5.2-.75 1.2-12.28-8.53-13.36-4.4-1.08 4.11 11.77 5.3 10.98 8.15-.8 2.85-9.06-5.38-10.74-2.18-1.7 3.21 11.65 6.98 11.76 7.01 4.3 1.12 15.25 3.49 19.08-2.12Z};\path[fill=hfcolorFF9D0B] svg {M77.4 48c1.62 0 3.07.66 4.07 1.87a5.97 5.97 0 0 1 1.33 3.76 7.1 7.1 0 0 1 1.95-.3c1.55 0 2.95.59 3.94 1.66a5.8 5.8 0 0 1 .8 7 5.3 5.3 0 0 1 1.78 2.82c.24.9.48 2.8-.8 4.74a5.22 5.22 0 0 1 .37 5.02c-1.02 2.32-3.57 4.14-8.51 6.1-3.08 1.22-5.9 2-5.92 2.01a44.33 44.33 0 0 1-10.93 1.6c-5.86 0-10.05-1.8-12.46-5.34-3.88-5.69-3.33-10.9 1.7-15.92 2.78-2.78 4.63-6.87 5.01-7.77.78-2.66 2.83-5.62 6.24-5.62a5.7 5.7 0 0 1 4.6 2.46c1-1.26 1.98-2.25 2.87-2.82A7.4 7.4 0 0 1 77.4 48Zm0 4c-.51 0-1.13.22-1.82.65-2.13 1.36-6.25 8.43-7.76 11.18a2.43 2.43 0 0 1-2.14 1.31c-1.54 0-2.75-1.53-.14-3.48 3.91-2.93 2.54-7.72.67-8.01a1.54 1.54 0 0 0-.24-.02c-1.7 0-2.45 2.93-2.45 2.93s-2.2 5.52-5.97 9.3c-3.78 3.77-3.98 6.8-1.22 10.83 1.87 2.75 5.47 3.58 9.15 3.58 3.82 0 7.73-.9 9.93-1.46.1-.03 13.45-3.8 11.76-7-.29-.54-.75-.76-1.34-.76-2.38 0-6.71 3.54-8.57 3.54-.42 0-.71-.17-.83-.6-.8-2.85 12.05-4.05 10.97-8.17-.19-.73-.7-1.02-1.44-1.02-3.14 0-10.2 5.53-11.68 5.53-.1 0-.19-.03-.23-.1-.74-1.2-.34-2.04 4.88-5.2 5.23-3.16 8.9-5.06 6.8-7.33-.23-.26-.57-.38-.98-.38-3.18 0-10.67 6.82-10.67 6.82s-2.02 2.1-3.24 2.1a.74.74 0 0 1-.68-.38c-.87-1.46 8.05-8.22 8.55-11.01.34-1.9-.24-2.85-1.31-2.85Z};\path[fill=hfcolorFFD21E] svg {M56.33 76.69c-2.75-4.04-2.56-7.07 1.22-10.84 3.77-3.77 5.97-9.3 5.97-9.3s.82-3.2 2.7-2.9c1.86.3 3.23 5.08-.68 8.01-3.92 2.93.78 4.92 2.28 2.17 1.51-2.75 5.63-9.82 7.76-11.18 2.13-1.35 3.64-.6 3.13 2.2-.5 2.79-9.42 9.55-8.55 11 .86 1.47 3.92-1.71 3.92-1.71s9.58-8.71 11.66-6.44c2.08 2.27-1.58 4.17-6.8 7.33-5.23 3.16-5.63 4-4.9 5.2.75 1.2 12.28-8.53 13.36-4.4 1.08 4.11-11.76 5.3-10.97 8.15.8 2.85 9.05-5.38 10.74-2.18 1.69 3.21-11.65 6.98-11.76 7.01-4.31 1.12-15.26 3.49-19.08-2.12Z};\end{scope}}}
\author{%
\begin{tabular}{c@{\hspace{3.5em}}c}
\textbf{Guanxu Yu} & \textbf{Yuhang Yao}\\[2pt]
{\small Independent Research} & {\small Carnegie Mellon University}\\[1pt]
{\small\href{mailto:guanxu.yu.sv@gmail.com}{\texttt{guanxu.yu.sv@gmail.com}}} &
{\small\href{mailto:yuhangya@alumni.cmu.edu}{\texttt{yuhangya@alumni.cmu.edu}}}
\end{tabular}\\[8pt]
{\footnotesize
\href{https://guanxuyu-sv.github.io/Visual-Jev/}{\websiteicon\, Website}
\quad\textcolor{lightgray}{\textbar}\quad
\href{https://github.com/guanxuyu-sv/Visual-Jev}{\githubicon\, GitHub}
\quad\textcolor{lightgray}{\textbar}\quad
\huggingfaceicon\, \textcolor{cvprblue}{Hugging Face:}
\href{https://huggingface.co/guanxuyu/visual-jev-4b-answer-sft}{visual-jev-4b}
\textcolor{gray}{/}
\href{https://huggingface.co/guanxuyu/visual-jev-8b-answer-sft}{visual-jev-8b}}}

\begin{document}
\raggedbottom
\maketitle

\begin{abstract}
Applications increasingly ask multiple closed-choice questions about the same
image, requiring accurate probabilities and efficient inference. We present
\textbf{Visual Jev}, a visual decision model combining LM-head candidate
readout, answer-supervised post-training, and independent question branches
over a shared visual prefix. On held-out test images from four benchmarks,
task SFT raises the 4B model's macro accuracy from \VJ{api.M0.accuracy.pct} to
\VJ{api.M1.accuracy.pct} and error-discrimination AUROC from
\VJ{rk.M0.macro.auroc} to \VJ{rk.M1.macro.auroc}. Its macro accuracy exceeds
the hosted Jev and OpenAI Decisions APIs by \VJ{api.gain_vs_jev.pp} and
\VJ{api.gain_vs_luna.pp} percentage points, respectively, with the lowest NLL.
We further show that better accuracy and calibration do not ensure reliable
decisions. Fixed confidence thresholds exceed target error rates, and
simply reordering options changes \VJ{perm.M1.snli_ve.flip@0.05.pct} of the
model's accept-or-review decisions on SNLI-VE.
At 32 questions per image, shared-prefix
inference reduces amortized time per question from \VJ{srv.S0.cold.32.msq}
to \VJ{srv.S1.cold.32.msq}~ms, a \VJ{srv.S1.speedup32}$\times$ speedup over
batched full sequences on one RTX 5090. Engine-level prefix caching reduces
KV memory to \VJ{srv.kvpercent} of that used by our custom shared-prefix path.
\end{abstract}

\section{Introduction}
\label{sec:intro}

A growing class of applications asks a vision--language model (VLM) to decide
rather than to write: which button cancels an order, whether a statement about
a photo is supported, how many seats are free. The candidate set is known at
request time, several questions often share one image, and the caller acts on
the returned probability by accepting confident answers and sending the rest
to a person. \Cref{fig:v3example} shows such a decision. A model returns
\VJ{ex.o0.top} for the correct option, which clears a threshold of
\VJ{ex.thr} fitted for 5\% risk, so the answer is accepted. When the same
options are listed in another order, the probability drops to \VJ{ex.o1.top}
and the question goes to review.

\input{generated/v3/fig_v3_example}

A deployment must satisfy several distinct requirements. Accuracy measures
how often an answer is correct. Calibration asks whether confidence agrees
with empirical correctness, while error ranking determines which answers can
be retained at low risk. Finally, a frozen threshold must be evaluated on new
images and under changes in presentation and execution. Good average
probability quality alone does not establish these deployment properties.

Prior work studies VLM calibration~\citep{tu2024empirical}, selective
VQA~\citep{whitehead2022reliable,eisenschlos2024selectively}, and sensitivity
to option permutations~\citep{zong2024fool}. We connect these perspectives
through a common operational endpoint: a threshold selected on calibration
images and subsequently held fixed. Our focus is whether improvements in
probability quality translate into useful coverage, and whether the resulting
accept-or-review decisions remain stable. An acceptance change can add review
work or admit an error; its frequency is a stability diagnostic, not itself a
measure of excess risk.

We develop a visual decision model that encodes an image and public context
once, executes independent question suffixes, and returns candidate
probabilities from the backbone's LM head. Answer-supervised post-training
learns the task, while temperature scaling (TS) and a frozen threshold map
its probabilities to accept-or-review decisions. Four benchmarks cover
relational questions, visual entailment, text recognition, and counting; the
latter two are excluded from post-training. We compare models before and
after training and calibration, then evaluate two hosted decision APIs on
the same splits to test whether deployment concerns extend beyond our model.

The study addresses three questions. \textbf{RQ1:} How do post-training and
calibration affect probability quality and frozen-threshold risk and coverage?
\textbf{RQ2:} How stable are decisions under option reordering, repeated API
calls, and changes in execution path? \textbf{RQ3:} What efficiency tradeoffs
arise when serving independent questions over shared visual context?
Matched training continuations and readout controls test alternatives to the
answer-SFT plus TS baseline.

We make three contributions:
\begin{enumerate}[nosep,leftmargin=*]
\item We introduce Visual Jev for answering independent closed-choice
      questions about a shared image. Across four benchmark test sets,
      answer-supervised post-training raises the 4B model's macro accuracy
      from \VJ{api.M0.accuracy.pct} to \VJ{api.M1.accuracy.pct} and
      error-discrimination AUROC from \VJ{rk.M0.macro.auroc} to
      \VJ{rk.M1.macro.auroc} (\Cref{sec:task,sec:quality}).
\item We show that better accuracy and calibration do not ensure reliable
      acceptance decisions. Our evaluation across four benchmarks and two
      hosted APIs identifies threshold failures and separately measures
      option-order sensitivity and variability across repeated requests
      (\Cref{sec:quality,sec:selective,sec:order,sec:hosted}).
\item We implement shared-prefix inference that is
      \VJ{srv.S1.speedup32}$\times$ faster than batched full sequences at
      32 questions per image. Comparisons on identical weights show that
      engine-level caching reduces KV memory to \VJ{srv.kvpercent} of our
      custom path's footprint, and quantify latency, sustained throughput,
      and decision stability (\Cref{sec:serving}).
\end{enumerate}

\section{Related work}
\label{sec:related}

We connect candidate readout, probability evaluation, and shared-context
serving through the same visual decision workload.

\paragraph{Vision--language models and how they are asked.}
Instruction-tuned VLMs~\citep{alayrac2022flamingo,li2023blip2,liu2023llava,dai2023instructblip}
are usually queried by generation, and the Qwen-VL line we build
on~\citep{bai2023qwenvl,wang2024qwen2vl,qwen3vl2025} follows that convention.
Reading option symbols directly is a strong way to query a language model on
multiple-choice problems~\citep{robinson2023leveraging}, and visual entailment
poses a typed three-way decision over an image and a
statement~\citep{xie2019visual,do2020snli}.

\paragraph{Decision-only models.}
TypeSafe's \textsc{Jev} answers closed questions with structured values instead
of text, trained for calibrated probabilities by reinforcement
learning~\citep{typesafejev2026}. The cited announcement describes text decisions; the API version evaluated
here also exposes experimental image input. We compare its returned
probabilities with those of an answer-supervised visual model and another
hosted decision API, without assuming access to their training procedures.

\paragraph{Calibration, scoring and selective prediction.}
Modern networks are often overconfident, and a single temperature fitted on
held-out data is a strong correction~\citep{guo2017calibration}; the same holds
for VLMs across distribution and label-set shifts~\citep{tu2024empirical}. We
score probabilities with the negative log likelihood (NLL) and the Brier
score~\citep{brier1950verification}, both strictly
proper~\citep{gneiting2007strictly}, and report expected calibration error
(ECE)~\citep{naeini2015obtaining} only as an auxiliary. Selective prediction
asks a model to abstain when unsure~\citep{elyaniv2010foundations,geifman2017selective};
in VQA, abstention has been studied as a route to
reliability~\citep{whitehead2022reliable,eisenschlos2024selectively}, and
language models have been probed for whether they know what they
know~\citep{kadavath2022language,rajpurkar2018know}. Because the area under the
risk--coverage curve (AURC) falls with accuracy alone, we also report its
excess over a perfect ranking~\citep{geifman2019bias} and an AUROC that compares confidence conditional on correctness~\citep{traub2024overcoming}. We evaluate the deployment
form, a threshold chosen on calibration data and frozen.

\paragraph{What candidate sets give away.}
Converting open answers to options invites shortcuts: hypothesis-only and
language-prior baselines expose
them~\citep{poliak2018hypothesis,gururangan2018annotation,goyal2017making,agrawal2018dont},
and option order is itself a bias for language
models~\citep{zheng2024large,pezeshkpour2024large} and for
VLMs~\citep{zong2024fool}. Those studies measure accuracy under reordering; we
measure what reordering does to a frozen accept-or-review decision.

\paragraph{Serving: caching and batching.}
Prefix reuse, paged key--value memory and continuous batching are the standard
levers for throughput~\citep{yu2022orca,kwon2023pagedattention,pope2023efficiently,zheng2024sglang};
kernel-level work reduces constant
factors~\citep{dao2022flashattention,shazeer2019fast,ainslie2023gqa}. We
compare a custom shared-prefix path against a production engine with
automatic prefix caching on the same weights, and measure numerical decision
drift under mixed-precision execution~\citep{micikevicius2018mixed}.

\section{Visual decision model}
\label{sec:task}

Visual Jev returns a candidate distribution for each independent question
about an image and public context. \Cref{fig:arch} shows how candidate
readout, post-training, and shared execution support the decision path.

\input{fig_architecture}

\subsection{Architecture overview}
\label{sec:architecture}

\paragraph{Shared visual context.} Visual Jev builds on Qwen3-VL's vision
tower, causal language decoder, and vocabulary projection (LM head).
Given an image $I$ and public text context $S$, the vision tower produces
visual-token embeddings. These are inserted into a prefix containing the
system instructions and $S$. Prefilling the decoder produces a per-layer
key--value (KV) cache that represents this common context. Qwen3-VL also
injects visual features into early decoder layers at image-token positions;
all such positions are in the prefix, so its cache captures their contribution.

\paragraph{Independent question branches.} Each question $q_i$ and its
candidates $\mathcal C_i$ form a separate suffix. All branches use the same
decoder weights and prefix state, while maintaining separate attention
contexts: a branch sees the image, public context and its own suffix.
Question-specific KV states are extended independently. This preserves the
single-question conditioning when several questions are evaluated together.

\paragraph{From hidden states to decisions.} The final hidden state at each
suffix's \texttt{Answer:} position is projected through the LM head. We select
the option-letter logits and return a distribution over that question's
candidates. Temperature scaling and a frozen acceptance cutoff operate on
this output as postprocessing stages (\Cref{fig:arch}); the same readout is
used with both Hugging Face Transformers (HF) execution and engine-level
prefix caching.

\subsection{Candidate probability readout}
\label{sec:readout}

\textbf{Choice} accepts a runtime-supplied set of
$K\le\VDMKmax{}$ candidates; \textbf{Claim} uses supported, contradicted and
not determined. Candidate $j$ maps to a verified single option-letter token.
For readout hidden state $h_i$, we gather its vocabulary logit
$z_j=(W_{\mathrm{LM}}h_i)_{a_j}$ at the option token $a_j$.
The returned distribution is
\begin{equation}
 p(c_j\mid I,S,q)=\frac{\exp(z_j/T)}{\sum_{\ell=1}^{K}\exp(z_\ell/T)},
 \label{eq:readout}
\end{equation}
where $T=1$ gives raw probabilities. This distribution is conditional on the
supplied candidates; it cannot detect a missing correct answer.

\paragraph{Why use the LM head?} A typed Choice/Claim head at the same
readout position is a matched alternative. On the same varied-$K$ training
corpus and budget, the LM head has higher macro accuracy and lower NLL and
Brier score after TS (\Cref{tab:v3design}(a); full contrasts in
Table~\ref{tab:v3contrastfull} in Appendix~\ref{sec:v3app}). We therefore use the backbone's existing head,
without additional classification parameters. This control uses A2k/H2k,
separately from the main M1 adapter.

\subsection{Answer supervision and calibration}
\label{sec:training}

We adapt the language tower with LoRA, freeze the vision tower, and supervise
the correct option-letter token $a_y$. For the decision-supervised training
items $\mathcal D$, answer SFT minimizes full-vocabulary next-token loss:
\begin{equation}
 \mathcal L_{\mathrm{SFT}}=-\frac{1}{|\mathcal D|}
 \sum_{(I,S,q,y)\in\mathcal D}
 \log P_\theta(a_y\mid I,S,q).
 \label{eq:sft}
\end{equation}
Training normalizes over the entire vocabulary; inference normalizes the
option logits over the supplied candidates as in \Cref{eq:readout}.
Only GQA and SNLI-VE supply SFT examples. Training batches complete
image--question examples independently; prefix reuse is applied at inference.
Training settings are in
Appendix~\ref{sec:repro}.

After training, TS fits one positive temperature per checkpoint and task by
minimizing calibration NLL. It preserves each answer's argmax. For a nominal
risk target $r$, we select the largest confidence-ranked calibration prefix
with error at most $r$, then freeze its last confidence as cutoff $\tau_r$.
An answer is accepted when $\max_j p(c_j)\ge\tau_r$, including all ties at the
cutoff. This empirical rule provides no finite-sample risk guarantee; risk
and coverage are measured separately on test images.

\subsection{Independent inference over shared context}
\label{sec:execution}

Our HF path prefills the prefix once and copies
its KV cache into separate question branches. Suffixes are left-padded and
batched, with padding masked from attention, so each row's final position is
its readout. Position identifiers are computed for each complete
prefix--suffix sequence and retained when the execution is split.
In vLLM, each question is an independent request; automatic prefix caching
lets requests share cached prefix blocks instead of materializing a prefix
copy per branch. Both paths reuse the image-conditioned prefix, while
their cache layouts have different memory costs (\Cref{sec:serving}).

\input{generated/v3/tab_v3_design}

\paragraph{Prefix sharing reduces computation.} With identical merged weights
and natural questions about each image, shared-prefix execution reduces time
per question from \VJ{srv.S0.cold.8.msq} to \VJ{srv.S1.cold.8.msq}~ms at
$N=8$ and from \VJ{srv.S0.cold.32.msq} to \VJ{srv.S1.cold.32.msq}~ms at
$N=32$ (\Cref{tab:v3design}(b)). \Cref{sec:serving} compares engine caching,
KV memory and sustained load; \Cref{sec:order} checks decision drift.

\section{Experimental setup}
\label{sec:setup}

\paragraph{Benchmarks and splits.} GQA supplies object, attribute and
relation questions; SNLI-VE supplies visual entailment. Both contribute SFT
training examples. TextVQA (text recognition, $K=8$) and TallyQA (counting,
$K=6$) are excluded from SFT. We convert open answers to candidate sets with
same-type distractors and shuffle positions. The main TallyQA build removes
images matched to GQA training; construction details and the separate short
subset are in Appendix~\ref{sec:data_build}.

SFT uses source-training splits; evaluation uses GQA source validation and
SNLI-VE source test, removing SNLI-VE images shared with our training set.
Each evaluation pool is split by parent-image hash into 30\% calibration and
70\% locked test, identically for all systems. Test sizes are
\VJ{n.gqa_val.test} GQA, \VJ{n.snli_ve.test} SNLI-VE,
\VJ{n.textvqa_k8.test} TextVQA and \VJ{n.tallyqa_l.test} TallyQA questions.
All benchmark scores use test images. Even tasks excluded from SFT supply
labeled calibration images; this evaluates within-task deployment, not
transfer to an uncalibrated task.

\paragraph{Systems and metrics.} M0 is Qwen3-VL-4B-Instruct~\citep{qwen3vl2025};
M1 adds answer-SFT LoRA~\citep{hu2022lora}, with three training seeds.
M2/M3 add task-local TS to M0/M1. Controls include matched LM/typed heads,
8B models and matched-budget continuations (Appendices~\ref{sec:repro} and~\ref{sec:d2}). Hosted
APIs receive no task-specific fine-tuning (Appendix~\ref{sec:api}).
We report accuracy, NLL, summed multiclass Brier score, auxiliary 15-bin ECE,
and AURC using maximum probability. E-AURC and AUROC assess error ranking
with accuracy discounted. Coverage is the fraction accepted by a frozen
cutoff; risk is the error fraction among accepted answers, undefined if none
are accepted. Macro averages give equal weight to four benchmarks. Intervals
are 95\% parent-image bootstrap intervals; multi-seed intervals average seeds
within each resample and capture image sampling, not training randomness.

\paragraph{Serving protocol.} We use one RTX 5090, merged SFT weights in
\texttt{bfloat16}, default image processing, and the same natural GQA questions
on each path. Timing includes preprocessing, transfer, execution and logit
readback from a decoded image. vLLM is version 0.30; full settings are in
Appendix~\ref{sec:repro}.

\section{RQ1: Accuracy and reliable acceptance}
\label{sec:results}

\paragraph{Answer to RQ1.} SFT improves accuracy, error ranking, and accepted
coverage; TS lowers NLL and ECE. Under the frozen cutoff rule, calibrated
probabilities still yield above-target error rates on held-out images.

\subsection{Answer and probability quality}

\label{sec:quality}

\input{generated/v3/tab_v3_quality_main}

\begin{figure*}[t]
\centering
\includegraphics[width=\textwidth]{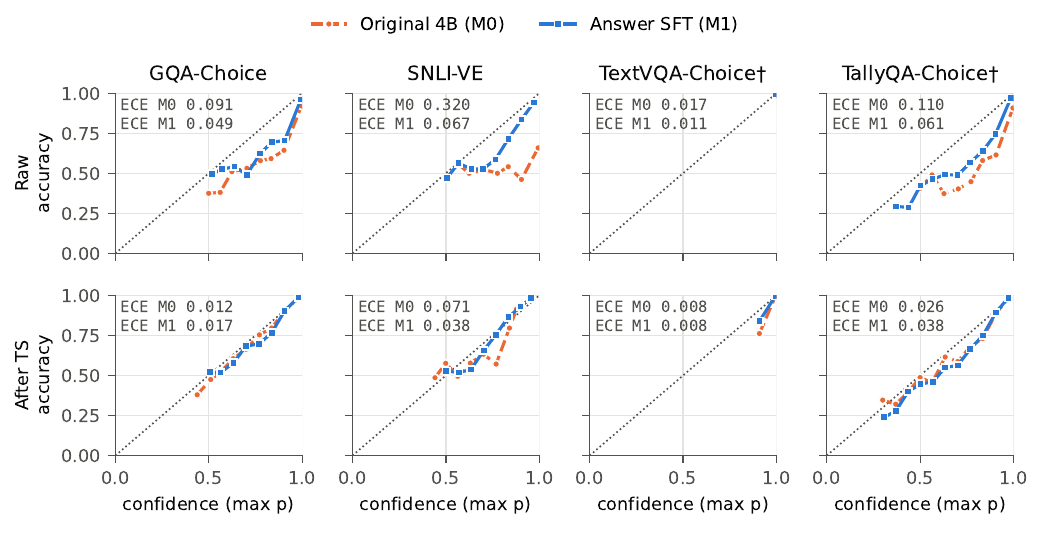}
\caption{TS reduces overall overconfidence; residual calibration error varies
by benchmark. Reliability on the test images, before (top) and after (bottom) task-local
TS; bins with fewer than 20 questions are omitted.}
\label{fig:v3rel}
\end{figure*}

\paragraph{Post-training improves accuracy on the trained families.}
Answer SFT raises macro accuracy by \VJ{d.M1-M0.macro.accuracy}
\VJ{d.M1-M0.macro.accuracy.ci} (Table~\ref{tab:v3qualitymain} and Table~\ref{tab:v3contrastfull} in Appendix~\ref{sec:v3app}). The gain
is concentrated on GQA (\VJ{M0.raw.gqa_val.accuracy} to
\VJ{M1.raw.gqa_val.accuracy}) and SNLI-VE (\VJ{M0.raw.snli_ve.accuracy} to
\VJ{M1.raw.snli_ve.accuracy}). TextVQA is near the ceiling for both, and on
held-out counting the post-trained model is worse
(\VJ{M1.raw.tallyqa_l.accuracy} against \VJ{M0.raw.tallyqa_l.accuracy}, lower
for every seed).

\paragraph{SFT improves error discrimination beyond its accuracy gain.} AURC falls
from \VJ{M0.raw.macro.aurc_maxp} to \VJ{M1.raw.macro.aurc_maxp}, but part of
that drop follows from higher accuracy. After task-local TS, AUROC rises from
\VJ{rk.M0.macro.auroc} to \VJ{rk.M1.macro.auroc}
(\VJ{rk.d.M1-M0.macro.auroc} \VJ{rk.d.M1-M0.macro.auroc.ci}), and the E-AURC change is \VJ{rk.d.M1-M0.macro.e_aurc} (Table~\ref{tab:v3ranking} in Appendix~\ref{sec:v3app}). On counting, where
accuracy drops, AUROC still rises from \VJ{rk.M0.tallyqa_l.auroc} to
\VJ{rk.M1.tallyqa_l.auroc}.

\paragraph{Both models are overconfident.} Fitted temperatures are
\VJ{T.M0.range} for the original backbone and \VJ{T.M1.range} after
post-training, all above one. Post-training removes part of the
overconfidence, lowering raw NLL from \VJ{M0.raw.macro.nll} to
\VJ{M1.raw.macro.nll}. Even so, \VJ{M1.raw.gqa_val.errconf09} of its GQA errors
are made with confidence of at least 0.9.

\paragraph{TS improves calibration; SFT improves discrimination.} TS lowers NLL to
\VJ{M0.ts_local.macro.nll} and \VJ{M1.ts_local.macro.nll} and ECE to
\VJ{M0.ts_local.macro.ece15} and \VJ{M1.ts_local.macro.ece15}, but barely moves
AURC (\VJ{d.M3-M1.macro.aurc_maxp.ci} for answer SFT,
\VJ{d.M2-M0.macro.aurc_maxp.ci} for the original model). One temperature keeps
each question's argmax; with more than two options it can reorder maximum
probabilities across questions; the measured AURC contrasts remain near zero. After TS, the ECE contrast is
\VJ{d.M3-M2.macro.ece15} \VJ{d.M3-M2.macro.ece15.ci}, with an interval
spanning zero; answer SFT's additional benefit is stronger error discrimination.
The 8B comparison shows the same broad pattern. \Cref{fig:v3rel} shows how TS moves confidence closer to observed accuracy across the four benchmarks.

The readout choice is validated in \Cref{sec:readout}; option-count and scale controls are in Appendix~\ref{sec:analysis}.

\paragraph{SFT plus TS leads the matched training controls.} It achieves
the lowest macro NLL and ECE after calibration across the tested
1,000-update continuations. Candidate CE, candidate Brier, and option
shuffling leave these metrics unimproved; continuing the original objective
controls for the extra updates (Appendix~\ref{sec:d2}).

\subsection{Risk and coverage}

\label{sec:selective}

\begin{figure*}[t]
\centering
\includegraphics[width=\textwidth]{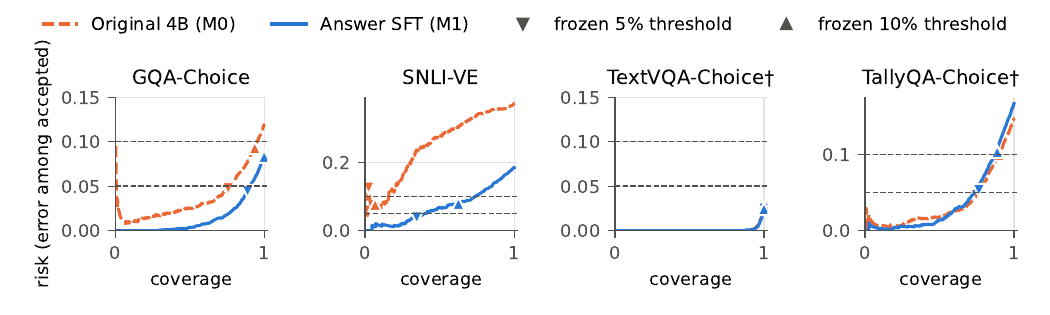}
\caption{Test risk--coverage curves and frozen deployment points. Answer SFT
increases coverage most clearly on the trained task families. Risk--coverage on the test images
after TS (one seed's curve per system), with the frozen 5\%
($\blacktriangledown$) and 10\% ($\blacktriangle$) points averaged over
seeds; dashed lines are the targets.}
\label{fig:v3rc}
\end{figure*}

\input{generated/v3/tab_v3_selective_main}

\paragraph{Frozen thresholds miss nominal risk targets.}
\Cref{tab:v3selectivemain} shows the test risk and coverage at a nominal
5\% target; Table~\ref{tab:v3selci} in Appendix~\ref{sec:v3app} also reports
the 10\% target. On TallyQA,
answer SFT's 5\% threshold yields
\VJ{M1.ts_local.tallyqa_l.risk@0.05.pct} risk
\VJ{M1.ts_local.tallyqa_l.risk@0.05.ci} at
\VJ{M1.ts_local.tallyqa_l.cov@0.05.pct} coverage; its 10\% operating point also
exceeds the target. A larger deviation occurs for the original model on
SNLI-VE: its 5\% threshold accepts only
\VJ{M0.ts_local.snli_ve.cov@0.05.pct}, with realized risk
\VJ{M0.ts_local.snli_ve.risk@0.05}
\VJ{M0.ts_local.snli_ve.risk@0.05.ci}. The wide interval reflects the small accepted set.

\paragraph{SFT expands coverage and reduces risk on SNLI-VE.} At the
nominal 5\% target, SFT raises coverage from
\VJ{M0.ts_local.snli_ve.cov@0.05.pct} to
\VJ{M1.ts_local.snli_ve.cov@0.05.pct} and reduces realized risk from
\VJ{M0.ts_local.snli_ve.risk@0.05.pct} to
\VJ{M1.ts_local.snli_ve.risk@0.05.pct}. AUROC rises from
\VJ{rk.M0.snli_ve.auroc} to \VJ{rk.M1.snli_ve.auroc}: the post-trained model
retains more correct answers while filtering errors more effectively.

\paragraph{Calibration depends on the task and evaluation build.} On the short
TallyQA subset, with \VJ{n.tallyqa_short.calib} calibration questions, the same
procedure misses its 5\% target (realized
\VJ{short.M1.ts_local.tallyqa.risk@0.05} for the post-trained model;
Table~\ref{tab:v3tally} in Appendix~\ref{sec:v3app}). The short build is count-balanced, whereas the larger build follows the
natural count distribution. Their accuracy differs substantially, so this
comparison does not isolate calibration sample size from distribution and
difficulty. A single temperature fitted on GQA and SNLI-VE calibration data
yields worse macro NLL and ECE than task-local TS across the four benchmarks
(Table~\ref{tab:v3transfer} in Appendix~\ref{sec:v3app}). \Cref{fig:v3rc} shows the risk--coverage curves and frozen operating points.

\subsection{Comparison with hosted decision APIs}

\label{sec:hosted}
\input{generated/v3/tab_v3_api_main}

We evaluate the OpenAI Decisions API~\citep{openaidecisions2026}
(GPT-6 Luna, accessed through OpenRouter; abbreviated OpenAI Decisions) and
TypeSafe's Jev on the same questions, option orders, and image-level splits. Both return option
probabilities rounded to two decimals. We apply task-local TS to these
probabilities. Quality and risk evaluation retain failed requests as uniform
distributions; stability analyses use successful responses. API versions,
failure handling, and probability preprocessing are detailed in
Appendix~\ref{sec:api}.

\paragraph{Visual Jev leads overall accuracy and error discrimination.}
Visual Jev achieves the highest macro accuracy and error-discrimination AUROC
among the evaluated systems (\Cref{tab:v3apimain}). SFT raises the same 4B
model's accuracy from \VJ{api.M0.accuracy.pct} to \VJ{api.M1.accuracy.pct}
and AUROC from \VJ{api.M0.auroc} to \VJ{api.M1.auroc}. Its accuracy exceeds
Jev by \VJ{api.gain_vs_jev.pp} percentage points and OpenAI Decisions by
\VJ{api.gain_vs_luna.pp}, with the lowest NLL. Gains concentrate on GQA and
SNLI-VE, included in our SFT; Jev leads on TextVQA, TallyQA and macro ECE.
API training data are unknown, so training conditions are not matched.

\paragraph{Calibration leaves above-target API risk.}
TS improves NLL and ECE for both APIs, yet the OpenAI model's frozen 5\% SNLI-VE
threshold yields risk \VJ{api.API_luna.snli_ve.risk@0.05}
\VJ{api.API_luna.snli_ve.risk@0.05.ci}. Its 10\% TallyQA threshold also
exceeds the target (\VJ{api.API_luna.tallyqa_l.risk@0.1}
\VJ{api.API_luna.tallyqa_l.risk@0.1.ci}). Both intervals lie above the nominal
targets on original-order test questions under the frozen cutoff rule.

\section{RQ2: Decision stability}
\label{sec:order}

\paragraph{Answer to RQ2.} Option reordering changes accept-or-review
decisions after SFT and TS. Hosted APIs expose two distinct failure modes:
sensitivity to presentation and variability across identical requests.
Backend changes also perturb frozen decisions.

\subsection{Option order and execution path}

\input{generated/v3/tab_v3_order}

\paragraph{Mean accuracy hides unstable decisions.} Reordering the options
leaves mean accuracy within about one point, but not the individual answers
(\Cref{tab:v3order}). For the original model, all four orders agree on
\VJ{perm.M0.snli_ve.consistent.pct} of SNLI-VE questions and only
\VJ{perm.M0.tallyqa.consistent.pct} of short-subset TallyQA questions, where
it picks the first displayed option \VJ{perm.M0.tallyqa.first_pos.pct} of the
time against \VJ{perm.M0.tallyqa.first_position_chance.pct} for a
position-blind model.

\paragraph{Answer agreement improves, but acceptance remains unstable.}
Agreement rises on every benchmark (SNLI-VE
\VJ{perm.M1.snli_ve.consistent.pct}). With the frozen 5\% threshold, however,
\VJ{perm.M1.snli_ve.flip@0.05.pct} of SNLI-VE questions are accepted in one
order and sent to review in another, as in \Cref{fig:v3example}. Averaging the
distribution over four orders costs four forward passes and leaves AURC at
\VJ{ens.M1.perm_avg_ts.aurc} against \VJ{ens.M1.single_ts.aurc}
(Table~\ref{tab:v3ens} in Appendix~\ref{sec:v3app}).

\paragraph{Execution changes also perturb frozen decisions.} With the same frozen
thresholds, moving between the four serving paths of \Cref{sec:serving}
changes the accept decision for \VJ{drift.matrix.flip5.min} to
\VJ{drift.matrix.flip5.max} of \VJ{drift.matrix.n} GQA questions, and moving
between HF and vLLM with or without prefix caching changes it for at most
\VJ{drift.flip5.maxpct} of a benchmark's calibration questions
(Table~\ref{tab:v3parity} in Appendix~\ref{sec:v3app}). Order flips count any disagreement among four presentations; backend flips
compare pairs of paths on separate samples. We report the diagnostics
separately. Even switching vLLM prefix caching changes numerical outputs.
A reproducible deployment therefore fixes both option order and cache settings.

\begin{figure*}[t]
\centering
\includegraphics[width=\textwidth]{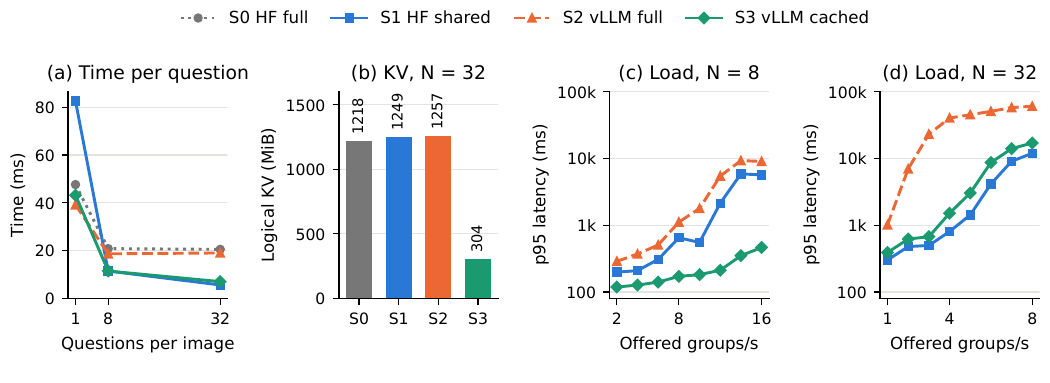}
\caption{Prefix reuse improves latency; engine caching also reduces memory.
(a) Cold per-question time, including image processing and prefill.
(b) Logical KV storage per image group at $N{=}32$, not peak GPU memory.
(c,d) p95 group latency under the same Poisson arrival schedules (200 groups
per offered rate), including queueing; logarithmic vertical axes.
All paths use the same weights on one RTX 5090. Full measurements are in
Tables~\ref{tab:v3serving} and~\ref{tab:v3load} in Appendix~\ref{sec:v3app}.}
\label{fig:v3serving}
\end{figure*}

\subsection{Reordered and repeated API requests}
\label{sec:api_stability}

\input{generated/v3/tab_v3_api_stability}

\paragraph{Repeated calls expose a second source of instability.}
\Cref{tab:v3apistability} reports the two diagnostics and their sample sizes.
Across the tested orders, \VJ{api.API_luna.perm.snli_ve.flip@0.05} of the OpenAI model's
SNLI-VE acceptance decisions disagree, compared with
\VJ{api.M1.perm.snli_ve.flip@0.05} for answer SFT. In our repeat check,
OpenAI Decisions returns identical rounded probabilities for every successfully
paired request. Jev instead changes its top answer on
\VJ{api.API_jev.repeat.snli_ve.changed} of SNLI-VE and
\VJ{api.API_jev.repeat.tallyqa.changed} of short-subset TallyQA questions.
Its disagreements across reordered requests therefore cannot be attributed
solely to option order. Answer changes under repetition and acceptance changes under reordering
measure distinct events on separate samples (\Cref{tab:v3apistability}).

OpenAI Decisions therefore demonstrates repeat consistency without order
robustness. A deployment audit must test both: replaying an unchanged
request checks repeat variability, whereas reordering candidates checks
whether presentation changes the accept-or-review action. SFT also expands
SNLI-VE acceptance coverage from \VJ{M0.ts_local.snli_ve.cov@0.05.pct} to
\VJ{M1.ts_local.snli_ve.cov@0.05.pct}. Its
\VJ{perm.M1.snli_ve.flip@0.05.pct} order-flip rate quantifies instability
at this higher coverage, complementing the gains in accuracy and useful
acceptance.

\Needspace{8\baselineskip}
\section{RQ3: Efficient inference over shared context}

\label{sec:serving}

\paragraph{Answer to RQ3.} At 32 questions per image, HF prefix reuse reduces
time per question from \VJ{srv.S0.cold.32.msq} to
\VJ{srv.S1.cold.32.msq}~ms, a \VJ{srv.S1.speedup32}$\times$ speedup
(\Cref{fig:v3serving}(a)). Engine-level sharing further lowers KV storage;
the fastest path under load depends on the number of questions per image.
We compare four paths on the same weights and images, first one group at a
time, then under sustained load.

\paragraph{Engine caching preserves the latency benefit.} The S0/S1 control
in \Cref{sec:execution} isolates HF prefix reuse. At $N=32$, vLLM without
prefix caching (S2) costs \VJ{srv.S2.cold.32.msq}~ms per question; caching
(S3) reduces it to \VJ{srv.S3.cold.32.msq}~ms, including image encoding and
prefill (\Cref{fig:v3serving}(a)). The custom HF shared path remains fastest for
this single-group workload.

\paragraph{The engine needs much less memory.} Engine caching reduces KV
storage (\Cref{fig:v3serving}(b)). S1 copies the prefix key--value state into
every question branch; S3 stores shared blocks once, so at $N{=}32$, a group
uses \VJ{srv.kvpercent} of S1's key--value memory and
\VJ{srv.capratio}$\times$ as many groups fit in a fixed budget. Over
\VJ{srv.agree.n} questions, S1, S2 and S3 change \VJ{srv.agree.S1.flips},
\VJ{srv.agree.S2.flips} and \VJ{srv.agree.S3.flips} argmax decisions relative
to S0.

\paragraph{Under load the winner depends on $N$.} We replay one Poisson
arrival schedule against each path (\Cref{fig:v3serving}(c,d)). With 8 questions per
image, vLLM with caching still keeps up at the highest rate we offered
(\VJ{load.S3.8.max} groups/s achieved). S1 serves one group at a time and
saturates at \VJ{load.S1.8.max}; at 8 groups/s its p95 latency is
\VJ{load.S1.8.8.p95}~ms against \VJ{load.S3.8.8.p95}~ms for S3. With 32
questions per image our path saturates later (\VJ{load.S1.32.max} against
\VJ{load.S3.32.max} groups/s) because it runs all suffixes in one padded
forward pass. Without prefix sharing the engine reaches lower throughput at $N{=}32$
(\VJ{load.S2.32.max} groups/s).

\paragraph{Native decision endpoint.} With the tested SFT seed, vLLM's
\texttt{/v1/systemone} prompt achieves macro accuracy
\VJ{s4.S4_a2_s0.acc}, against \VJ{s4.a2_s0.acc} for our readout
(Table~\ref{tab:v3s4} in Appendix~\ref{sec:v3app}).

\section{Conclusion}
Visual Jev combines accurate candidate probabilities with efficient inference
for independent questions over a shared image. Answer SFT raises macro
accuracy from \VJ{api.M0.accuracy.pct} to \VJ{api.M1.accuracy.pct} and
error-discrimination AUROC from \VJ{rk.M0.macro.auroc} to
\VJ{rk.M1.macro.auroc}, exceeding both evaluated hosted APIs in macro
accuracy and AUROC. Task-local temperature scaling improves probability
quality, and stronger error ranking expands useful acceptance coverage.

Our evaluation also exposes failures hidden by aggregate scores. Frozen
cutoffs exceed nominal error targets after calibration; option reordering
changes \VJ{perm.M1.snli_ve.flip@0.05.pct} of SNLI-VE accept-or-review
decisions; repeated API requests reveal a separate source of answer
variability. These results establish decision stability as an explicit
criterion alongside accuracy and calibration.

Shared-prefix execution delivers a \VJ{srv.S1.speedup32}$\times$ speedup at
32 questions per image, and engine caching reduces KV memory to
\VJ{srv.kvpercent} of the custom shared path's footprint. Under load, the
faster path depends on question count. Visual Jev thus links model adaptation,
decision reliability, and shared-context serving.
\mbox{See Appendix~\ref{sec:limitations} for limitations and scope.}

\section*{LLM Usage Statement}
\label{sec:llm}
The authors conceived the research idea, posed the research questions,
designed the experiment plan and directed the work throughout. We used Claude
Code (Anthropic, Claude Opus 5.5) to implement it. It wrote the code for
evaluation, training, analysis and serving benchmarks, ran the experiments,
generated the tables and figures, and suggested some of the analyses. It rewrote and revised the text, starting
from an earlier draft by the authors, and checked the references. The authors
verified the results, claims and references and take full responsibility for
the content. OpenAI Codex assisted with a subsequent structural and
editorial revision, claim--evidence checks, and figure and layout preparation.\par

{
    \fontsize{8.5}{10}\selectfont
    \setlength{\bibsep}{0pt}
    \bibliographystyle{ieeenat_fullname}
    \bibliography{refs}
}

\vfill\null
\clearpage
\maketitlesupplementary
\appendix
\fontsize{9.5}{11.2}\selectfont
\makeatletter\setlength{\@fptop}{0pt}\setlength{\@fpbot}{0pt plus 1fil}\setlength{\@dblfptop}{0pt}\setlength{\@dblfpbot}{0pt plus 1fil}\makeatother

\section{Supplementary results}
\label{sec:v3app}

\input{generated/v3/tab_v3_quality}
\input{generated/v3/tab_v3_ranking}
\input{generated/v3/tab_v3_transfer}
\input{generated/v3/tab_v3_serving}
\input{generated/v3/tab_v3_load}

\input{generated/v3/tab_v3_selective_ci}
\input{generated/v3/tab_v3_tally}
\input{generated/v3/tab_v3_contrast_full}
\input{generated/v3/tab_v3_byk}
\input{generated/v3/tab_v3_ensemble}
\input{generated/v3/tab_v3_parity}
\input{generated/v3/tab_v3_s4}

\paragraph{vLLM parity.} \Cref{tab:v3parity} checks the engine against HF on
the same prompts and weights before any timing is compared: with the merged
answer-SFT weights, \VJ{par.a2_s0.apc0.fliprate} of argmax decisions differ
and accuracy differs by \VJ{par.a2_s0.apc0.dacc} points. The comparison quantifies numerical differences between the two execution
paths; output extraction uses candidate logits on both sides.

\section{Data and ethics}
\label{sec:data_build}

\paragraph{Counting benchmark construction.} We build the counting
task from two shards of the full TallyQA test release, which keep its natural
count distribution and several questions per image. Its images carry no
identifiers, so we compare every image by size and a 64-bit difference hash
against the GQA training images and remove \VJ{n.tallyqa_l.dropimg} matches
(\VJ{n.tallyqa_l.dropq} questions), leaving \VJ{n.tallyqa_l.all} questions on
\VJ{n.tallyqa_l.images} images. We also report a count-balanced 498-question
subset separately (\Cref{tab:v3tally}): the same check
finds that \VJ{tallyqa_short.overlap} of its images
(\VJ{tallyqa_short.overlap_pct}) are copies of GQA training images. Accuracy
on those images is not higher (\VJ{tallyqa_short.M1.acc_overlap} against
\VJ{tallyqa_short.M1.acc_clean} for the post-trained model), so the observed comparison shows no overlap-associated accuracy gain.
The main counting build excludes all matched images.

This work uses four publicly released benchmarks, GQA~\citep{hudson2019gqa},
SNLI-VE~\citep{xie2019visual}, TextVQA~\citep{singh2019towards} and
TallyQA~\citep{acharya2019tallyqa}, under their original terms, together with
the images they are built on~\citep{krishna2017visual,young2014image}. The
images include identifiable people; apart from the one example image in
\Cref{fig:v3example} we redistribute none of them, and will release only
derived question records, option sets, model outputs and code. No
new human annotation was collected.

The intended use is a serving pattern, not a decision procedure with
consequences for people. Our results show what acting on its probabilities
requires: raw probabilities can be overconfident, task-local calibration
improves their quality, empirical thresholds are not guarantees, and option
order can change individual decisions.

\section{Reproduction details}
\label{sec:repro}
\input{txt_repro_cvpr}

\section{Commercial decision APIs}
\label{sec:api}
\input{generated/v3/tab_v3_api}

We ran the same protocol on two hosted decision APIs that return a probability
for every option: OpenAI Decisions (GPT-6 Luna; \texttt{openai/gpt-6-luna-decisions-20261006},
through OpenRouter's Decisions API) and TypeSafe's Jev
(\texttt{jev-1.14.0}, System One API, whose image input the provider marks as
not yet production quality). Each request carries one image and all of its
questions, with the options in the same displayed order as for our models;
temperatures and thresholds are fitted on the same calibration images.
Both APIs round probabilities to two decimals. We floor them at 0.005,
take logs, and renormalize before scoring or fitting TS. A floor of $10^{-4}$
preserves the accuracy ranking and the two above-target intervals in
\Cref{sec:hosted}, but also puts the OpenAI model's 5\% TallyQA interval above
its target. Failed requests (refusals for OpenAI Decisions, rejected image sizes for Jev) are kept as
uniform distributions: \VJ{api.API_luna.failed} and \VJ{api.API_jev.failed} of
questions. OpenRouter reported \$\VJ{api.API_luna.cost} for API requests;
TypeSafe reported no API charge.

Reordering statistics require a successful original response and at least
one successful reordering, retaining up to three reorderings per question.
They do not require complete four-order sets. Repeat statistics use only
successfully paired responses.

\paragraph{Accuracy and ranking.} Visual Jev (trained) leads macro accuracy
and AUROC (\cref{tab:v3api}). API-minus-M1 accuracy is
\VJ{api.d.jev_m1.macro.accuracy} \VJ{api.d.jev_m1.macro.accuracy.ci} for Jev
and \VJ{api.d.luna_m1.macro.accuracy} \VJ{api.d.luna_m1.macro.accuracy.ci}
for OpenAI Decisions. The lead is on held-out GQA and SNLI-VE, included in our
SFT. Both APIs beat M0 on SNLI-VE; M1 reverses this (\cref{tab:v3api}). We do
not fine-tune the APIs; their training data are unknown. Jev leads on the
excluded TextVQA and TallyQA tasks; short-TallyQA
accuracy is \VJ{api.API_jev.short.accuracy} versus \VJ{api.M1.short.accuracy}.

\paragraph{Calibration and thresholds.} Task-local TS lowers NLL and ECE for
both APIs. Unlike the OpenAI model (\Cref{sec:hosted}), all Jev risk intervals
include or lie below their nominal targets under the same cutoff rule.
At the nominal 5\% target, answer SFT accepts \VJ{api.M1.snli_ve.cov@0.05} of SNLI-VE questions,
OpenAI Decisions \VJ{api.API_luna.snli_ve.cov@0.05} and Jev \VJ{api.API_jev.snli_ve.cov@0.05}.
The APIs' probabilities are coarse: \VJ{api.API_luna.round.gqa_val.one} (OpenAI Decisions)
and \VJ{api.API_jev.round.gqa_val.one} (Jev) of GQA answers come back as
1.00.

\paragraph{Option order and repeated requests.} OpenAI Decisions returns
identical rounded distributions in successful repeat pairs. Its answer
agreement across available original and reordered responses is
\VJ{api.API_luna.perm.snli_ve.consistent} on SNLI-VE and
\VJ{api.API_luna.perm.tallyqa.consistent} on short TallyQA. Jev changes its top
answer on \VJ{api.API_jev.repeat.gqa_val.changed} of GQA,
\VJ{api.API_jev.repeat.snli_ve.changed} of SNLI-VE, and
\VJ{api.API_jev.repeat.tallyqa.changed} of short TallyQA repeat pairs. Its
\VJ{api.API_jev.perm.snli_ve.flip@0.05} SNLI-VE acceptance disagreement
reflects a setting with both order sensitivity and repeat variability;
the two diagnostics are reported separately (\Cref{tab:v3apistability}).

\section{Probability-targeted training controls}
\label{sec:d2}

\input{generated/v3/tab_v3_d2}

We continued each released seed under four objectives (\Cref{tab:v3d2}): its
own full-vocabulary cross entropy (CE), which controls for the extra training;
candidate CE and candidate Brier, which score the distribution the system
actually returns; and candidate CE with options reshuffled in every example.
At this budget and with no tuning, none improves on post-training plus TS.

\paragraph{Extra training alone hurts raw calibration.} Continuing the
original objective makes raw probabilities more overconfident (NLL
\VJ{D_full.raw.macro.nll} against \VJ{M1.raw.macro.nll}) and increases NLL after TS (\VJ{d.Dfull-M3.macro.nll} \VJ{d.Dfull-M3.macro.nll.ci}).
Among the continuations, Brier achieves the lowest raw NLL (\VJ{D_brier.raw.macro.nll}), but it still does not outperform the uncontinued SFT model after TS.

\paragraph{Shuffling improves answer agreement without stabilizing acceptance.} Option shuffling raises
SNLI-VE order agreement to \VJ{perm.D_ce_perm.snli_ve.consistent.pct} from
\VJ{perm.D_ce.snli_ve.consistent.pct} without it, but leaves the accept flip
rate near \VJ{perm.D_ce_perm.snli_ve.flip@0.05.pct}. Ranking metrics remain close across the arms (AUROC \VJ{rk.D_ceperm.macro.auroc} to
\VJ{rk.D_brier.macro.auroc}; \Cref{tab:v3ranking}). Within this budget, none of the tested objectives improves on the existing
SFT plus TS baseline. The tested schedule and training scope are specified
in Appendix~\ref{sec:limitations}.

\section{Readout, option count, and scale}
\label{sec:analysis}

The readout control in \Cref{sec:readout} is complemented by option-count
and model-scale analyses within the local backbone family.

\paragraph{Readout control.} The matched LM-head/typed-head comparison is
summarized in \Cref{sec:readout,tab:v3design}; paired intervals remain in
\Cref{tab:v3contrastfull}.

\paragraph{Option count.} Overconfidence appears at each tested option count,
and one temperature per benchmark leaves no systematic dependence on $K$
(\Cref{tab:v3byk}). GQA post-training uses $K\in\{2,4\}$, yet accuracy also
improves on GQA questions with $K{=}8$; SNLI-VE training uses three options.

\paragraph{Scale.} The 8B backbone improves the reported macro metrics before and after
post-training (\Cref{tab:v3quality,tab:v3ranking}), and its probabilities
follow the same broad pattern: overconfident raw, better calibrated after TS,
and stronger ranking after post-training.

\section{Limitations and scope}
\label{sec:limitations}
\input{txt_limitations_cvpr}

\end{document}

%% file: generated/macros.tex
\usepackage{xspace}
\newcommand{\VDMKmax}{16\xspace}

\newcommand{\VDMgqaBlind}{58.5\%\xspace}

\newcommand{\VDMgqaChance}{36.5\%\xspace}

\newcommand{\VDMsnliBlind}{33.4\%\xspace}

\newcommand{\VDMsnliChance}{33.3\%\xspace}

\newcommand{\VDMsteps}{3,000\xspace}

%% file: generated/v3/fig_v3_example.tex
\begin{figure}[t]
\centering
\small
\fbox{\parbox{0.96\columnwidth}{%
\begin{minipage}[t]{0.36\linewidth}\vspace{0pt}\includegraphics[width=\linewidth]{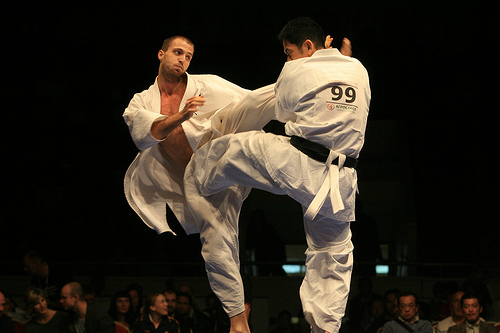}\end{minipage}\hfill
\begin{minipage}[t]{0.61\linewidth}\vspace{0pt}\footnotesize
\textbf{Statement:} ``Two men enjoy doing karate.''\\
\textit{Gold: not determined by the image}\\[2pt]
\begin{tabular}{@{}l@{\hspace{4pt}}l@{\hspace{3pt}}r@{}}
\multicolumn{3}{@{}l}{\textit{Original order}: top 0.89, \textbf{accept}} \\
& A.\ supported by the image & 0.08 \\
& B.\ contradicted by the image & 0.03 \\
& C.\ not determined by the image & 0.89 \\
\multicolumn{3}{@{}l}{\textit{Reordered}: top 0.78, \textbf{review}} \\
& A.\ contradicted by the image & 0.05 \\
& B.\ not determined by the image & 0.78 \\
& C.\ supported by the image & 0.16 \\
\end{tabular}\end{minipage}}}
\caption{Option-order sensitivity of selective prediction. For the answer-SFT model (seed 0), reordering the candidates in an SNLI-VE test example leaves the predicted answer unchanged but reduces its calibrated probability from 0.89 to 0.78, changing the decision from acceptance to review. The temperature and acceptance threshold (0.88) are fitted on calibration images and held fixed across orders; the threshold is selected for a target risk of 5\%.}
\label{fig:v3example}
\end{figure}

%% file: fig_architecture.tex
\begin{figure*}[t]
\centering
\resizebox{\textwidth}{!}{%
\begin{tikzpicture}[
  font=\small,
  src/.style    ={draw=vdmblue!55, fill=vdmblue!8, rounded corners=2pt,
                  align=center, inner sep=3pt, minimum height=7mm, minimum width=15mm},
  core/.style   ={draw=vdmblue!75, fill=vdmblue!16, rounded corners=2pt,
                  align=center, inner sep=4pt, minimum height=13mm, minimum width=22mm,
                  line width=0.7pt},
  branch/.style ={draw=black!25, fill=white, rounded corners=2pt,
                  align=center, inner sep=3pt, minimum height=6.5mm, minimum width=50mm},
  stage/.style  ={draw=vdmaqua!70, fill=vdmaqua!10, rounded corners=2pt,
                  align=center, inner sep=3pt, minimum height=13mm, minimum width=19mm},
  gate/.style   ={draw=black!45, fill=white, rounded corners=2pt,
                  align=center, inner sep=3pt, minimum height=13mm, minimum width=19mm},
  accept/.style ={draw=vdmaqua!80, fill=vdmaqua!22, rounded corners=2pt,
                  align=center, inner sep=2pt, minimum height=6mm, minimum width=15mm},
  review/.style ={draw=vdmorange!80, fill=vdmorange!16, rounded corners=2pt,
                  align=center, inner sep=2pt, minimum height=6mm, minimum width=15mm},
  flow/.style   ={-{Stealth[length=1.6mm,width=1.3mm]}, draw=black!55, line width=0.5pt},
  bus/.style    ={draw=black!55, line width=0.5pt},
  panel/.style  ={rounded corners=3pt, draw=black!14, line width=0.5pt},
  ptitle/.style ={font=\scriptsize\bfseries, text=black!45},
  rq/.style     ={font=\scriptsize\bfseries, text=white, fill=black!60,
                  rounded corners=1.5pt, inner xsep=2.5pt, inner ysep=1.2pt},
  sub/.style    ={font=\scriptsize, text=black!60},
]

\node[src]  (img) at (0, 0.55) {image};
\node[src]  (ctx) at (0,-0.55) {context};
\node[core] (prefix) at (2.75,0) {\textbf{shared prefix}\\[1pt]
  \scriptsize encoded once\\[-1pt]\scriptsize KV cache};
\draw[flow] (img.east) -- (img.east -| prefix.west);
\draw[flow] (ctx.east) -- (ctx.east -| prefix.west);

\node[branch] (q1) at (7.35, 0.78) {question 1 + options $\rightarrow$ \texttt{Answer:}};
\node[branch] (q2) at (7.35, 0) {question 2 + options $\rightarrow$ \texttt{Answer:}};
\node[text=black!70] (dots) at (7.35,-0.47) {$\vdots$};
\node[branch] (qn) at (7.35,-0.95) {question $N$ + options $\rightarrow$ \texttt{Answer:}};
\coordinate (split) at (4.45,0);
\draw[bus] (prefix.east) -- (split);
\draw[bus] (split |- q1) -- (split |- qn);
\foreach \t in {q1,q2,qn}{\draw[flow] (split |- \t) -- (\t.west);}

\coordinate (join) at (10.25,0);
\foreach \t in {q1,q2,qn}{\draw[bus] (\t.east) -- (join |- \t);}
\draw[bus] (join |- q1) -- (join |- qn);

\node[stage] (dist) at (11.7,0) {};
\node[font=\scriptsize\bfseries, anchor=north] at ([yshift=-1.5pt]dist.north) {LM-head $p$};
\foreach \x/\h in {-0.42/0.10, 0/0.05, 0.42/0.62}{
  \fill[vdmaqua!75] ($(dist.south)+(\x-0.13,0.12)$) rectangle ++(0.26,\h);}
\draw[black!40, line width=0.4pt] ($(dist.south)+(-0.68,0.12)$) -- ++(1.36,0);
\draw[flow] (join) -- (dist.west);

\node[stage] (ts) at (13.85,0) {\textbf{temperature}\\[1pt]$p \propto e^{z/T}$\\[-1pt]
  \scriptsize $T$ fit on calib.};
\draw[flow] (dist) -- (ts);

\node[gate] (thr) at (16.0,0) {\textbf{frozen}\\\textbf{threshold}\\[1pt]
  \scriptsize $\max p \ge \tau_r$\,?};
\draw[flow] (ts) -- (thr);

\node[accept] (acc) at (18.1, 0.42) {\scriptsize\bfseries accept};
\node[review] (rev) at (18.1,-0.42) {\scriptsize\bfseries review};
\draw[flow] (thr.east |- acc) -- (acc.west);
\draw[flow] (thr.east |- rev) -- (rev.west);

\path (0,1.65) coordinate (topA) (0,-1.78) coordinate (botA);
\begin{scope}[on background layer]
  \node[panel, fill=vdmblue!5, fit=(img)(ctx)(prefix)(topA)(botA),
        inner xsep=6pt, inner ysep=2pt] (pA) {};
  \node[panel, fill=black!3, fit=(q1)(qn)(q1 |- topA)(qn |- botA),
        inner xsep=6pt, inner ysep=2pt] (pB) {};
  \node[panel, fill=vdmaqua!4, fit=(dist)(ts)(thr)(acc)(rev)(dist |- topA)(dist |- botA),
        inner xsep=6pt, inner ysep=2pt] (pC) {};
\end{scope}
\node[ptitle, anchor=north west] at ([shift={(3pt,-3pt)}]pA.north west) {SHARED CONTEXT};
\node[ptitle, anchor=north west] at ([shift={(3pt,-3pt)}]pB.north west) {ISOLATED QUESTIONS};
\node[ptitle, anchor=north west] at ([shift={(3pt,-3pt)}]pC.north west) {DECISION POLICY};

\node[rq, anchor=south] at ([yshift=2pt]dist.north) {RQ1 ranking};
\node[rq, anchor=south] at ([yshift=2pt]ts.north) {RQ1 calibration};
\node[rq, anchor=south] at ([yshift=2pt]thr.north) {RQ1 risk};
\node[rq, anchor=south east] at ([shift={(-3pt,3pt)}]pB.south east) {RQ2 option order};
\node[rq, anchor=south east] at ([shift={(-3pt,3pt)}]pA.south east) {RQ2/3 backend};

\end{tikzpicture}}
\caption{Visual decision model and evaluation. A shared prefix (image and public context,
encoded once) serves isolated question suffixes, executed by HF Transformers or
by vLLM; the LM head's option-letter logits give the candidate distribution of
\cref{eq:readout}. The decision path on the right, temperature scaling and a
threshold fixed on calibration images, is what the research questions probe.
A typed-head variant is a readout control (\cref{sec:readout}).}
\label{fig:arch}
\end{figure*}

%% file: generated/v3/tab_v3_design.tex
\begin{table}[t]
\centering
\footnotesize
\setlength{\tabcolsep}{4pt}
\begin{adjustbox}{max width=\linewidth}
\begin{tabular}{lccc}
\toprule
\multicolumn{4}{l}{\textit{(a) Matched readouts: four-task macro}} \\
Readout & Acc.$\uparrow$ & NLL$\downarrow$ & Brier$\downarrow$ \\
\midrule
LM head & 0.889 & 0.304 & 0.161 \\
Typed head & 0.884 & 0.332 & 0.170 \\
\midrule
\multicolumn{4}{l}{\textit{(b) Same-weight HF execution: ms/question}} \\
Path & $N{=}8$ & $N{=}32$ & Speedup ($N{=}32$) \\
\midrule
Full sequences & 20.8 & 20.5 & 1.0$\times$ \\
Shared prefix & 11.4 & 5.6 & 3.7$\times$ \\
\bottomrule
\end{tabular}
\end{adjustbox}
\caption{Design controls on held-out data. (a) LM and typed heads use the same varied-$K$ training corpus and budget, three seeds, and task-local TS. These are A2k/H2k, separate from the released M1. (b) Same merged M1 weights, natural GQA questions and one RTX 5090; timing includes image processing and prefill. Protocols in \cref{sec:setup} and Appendix~\ref{sec:repro}.}
\label{tab:v3design}
\end{table}

%% file: generated/v3/tab_v3_quality_main.tex
\begin{table*}[t]
\centering
\small
\setlength{\tabcolsep}{4pt}
\begin{adjustbox}{max width=\linewidth}
\begin{tabular}{lccccccccc}
\toprule
 & \multicolumn{4}{c}{Accuracy} & \multicolumn{5}{c}{Macro over the four benchmarks} \\
\cmidrule(lr){2-5}\cmidrule(lr){6-10}
System & GQA & SNLI-VE & TextVQA$^\dagger$ & TallyQA$^\dagger$ & Acc$\uparrow$ & NLL$\downarrow$ & Brier$\downarrow$ & ECE$\downarrow$ & AURC$\downarrow$ \\
\midrule
Original 4B (M0) & 0.880 & 0.629 & 0.974 & 0.853 & 0.834 & 1.009 & 0.294 & 0.134 & 0.079 \\
\quad + TS (M2) & 0.880 & 0.629 & 0.974 & 0.853 & 0.834 & 0.416 & 0.230 & 0.029 & 0.080 \\
Answer SFT (M1) & 0.916 & 0.808 & 0.975 & 0.841 & 0.885 & 0.331 & 0.172 & 0.047 & 0.030 \\
\quad + TS (M3) & 0.916 & 0.808 & 0.975 & 0.841 & 0.885 & 0.309 & 0.165 & 0.025 & 0.030 \\
\bottomrule
\end{tabular}
\end{adjustbox}
\caption{Answer SFT improves accuracy and probability quality; TS further improves calibration. All scores use held-out test images; macro averages weight four benchmarks equally. SFT rows average three seeds. $^\dagger$Task excluded from SFT. The 8B results are in Table~\ref{tab:v3quality} in Appendix~\ref{sec:v3app}.}
\label{tab:v3qualitymain}
\end{table*}

%% file: generated/v3/tab_v3_selective_main.tex
\begin{table*}[t]
\centering
\small
\setlength{\tabcolsep}{4pt}
\begin{adjustbox}{max width=\linewidth}
\begin{tabular}{lcccc}
\toprule
 & \multicolumn{2}{c}{Original + TS (M2)} & \multicolumn{2}{c}{Answer SFT + TS (M3)} \\
\cmidrule(lr){2-3}\cmidrule(lr){4-5}
Benchmark & Coverage & Risk [95\% CI] & Coverage & Risk [95\% CI] \\
\midrule
GQA & 0.758 & 0.047 [0.041, 0.054] & 0.888 & 0.044 [0.039, 0.050] \\
SNLI-VE & 0.023 & 0.125 [0.029, 0.259] & 0.346 & 0.038 [0.025, 0.054] \\
TextVQA$^\dagger$ & 1.000 & 0.026 [0.019, 0.034] & 1.000 & 0.025 [0.017, 0.031] \\
TallyQA$^\dagger$ & 0.748 & 0.047 [0.039, 0.055] & 0.764 & 0.053 [0.046, 0.061] \\
\bottomrule
\end{tabular}
\end{adjustbox}
\caption{Frozen deployment at a nominal 5\% risk target, using a cutoff selected on ranked calibration answers. Coverage and risk are measured on test images. Brackets give 95\% image-cluster bootstrap intervals, not risk guarantees. Both 5\% and 10\% operating points appear in Table~\ref{tab:v3selci} in Appendix~\ref{sec:v3app}. $^\dagger$Task excluded from our post-training.}
\label{tab:v3selectivemain}
\end{table*}

%% file: generated/v3/tab_v3_api_main.tex
\begin{table}[t]
\centering
\small
\setlength{\tabcolsep}{4pt}
\begin{adjustbox}{max width=\linewidth}
\begin{tabular}{lcccc}
\toprule
System & Macro Acc.$\uparrow$ & AUROC$\uparrow$ & NLL$\downarrow$ & ECE$\downarrow$ \\
\midrule
Visual Jev (untrained) & 83.4\% & 0.834 & 0.416 & 0.029 \\
\textbf{Visual Jev (trained)} & \textbf{88.5\%} & \textbf{0.890} & \textbf{0.309} & 0.025 \\
Jev API & 85.8\% & 0.870 & 0.361 & \textbf{0.013} \\
OpenAI Decisions & 83.5\% & 0.833 & 0.440 & 0.020 \\
\bottomrule
\end{tabular}
\end{adjustbox}
\caption{Macro performance on the same held-out test questions, with equal weight for four benchmarks; column bests are bold. Trained/untrained = M1/M0, with/without our task SFT of the pretrained 4B. NLL, ECE and AUROC are after task-local TS. Failed API requests (Jev 1.9\%, OpenAI 0.3\%) receive uniform probabilities. API training data are unknown. Per-benchmark results are in Table~\ref{tab:v3api} in Appendix~\ref{sec:api}.}
\label{tab:v3apimain}
\end{table}

%% file: generated/v3/tab_v3_order.tex
\begin{table}[t]
\centering
\footnotesize
\setlength{\tabcolsep}{4pt}
\begin{adjustbox}{max width=\linewidth}
\begin{tabular}{lcccccc}
\toprule
 & \multicolumn{2}{c}{Same answer} & \multicolumn{2}{c}{Accept flip @5\%} & \multicolumn{2}{c}{First position} \\
\cmidrule(lr){2-3}\cmidrule(lr){4-5}\cmidrule(lr){6-7}
Benchmark & M0 & M1 & M0 & M1 & M0 & M1 \\
\midrule
GQA & 0.933 & 0.975 & 0.140 & 0.048 & 0.402 & 0.375 \\
SNLI-VE & 0.748 & 0.926 & 0.026 & 0.170 & 0.378 & 0.313 \\
TextVQA$^\dagger$ & 0.969 & 0.972 & 0.000 & 0.001 & 0.125 & 0.123 \\
TallyQA$^\dagger$ (short) & 0.325 & 0.464 & 0.086 & 0.078 & 0.315 & 0.248 \\
\bottomrule
\end{tabular}
\end{adjustbox}
\caption{Reordering changes individual answers and frozen decisions. SFT improves answer agreement, but does not uniformly reduce acceptance changes. Each test question is scored in its original order and three fixed reorderings, mapped back to semantic labels, with the frozen temperature and threshold. Same answer: all four orders agree. Accept flip: the 5\% threshold accepts one order and sends another to review. First position: how often the first displayed option is chosen.}
\label{tab:v3order}
\end{table}

%% file: generated/v3/tab_v3_api_stability.tex
\begin{table}[t]
\centering
\footnotesize
\setlength{\tabcolsep}{4pt}
\begin{adjustbox}{max width=\linewidth}
\begin{tabular}{lcccc}
\toprule
 & \multicolumn{2}{c}{Option reordering} & \multicolumn{2}{c}{Two unchanged calls} \\
\cmidrule(lr){2-3}\cmidrule(lr){4-5}
System & $n$ & Accept flip & $n$ & Answer change \\
\midrule
Answer SFT & 1,368 & 17.0\% & -- & -- \\
OpenAI Decisions & 1,368 & 17.5\% & 1,500 & 0.0\% \\
Jev API & 1,359 & 11.0\% & 1,492 & 2.7\% \\
\bottomrule
\end{tabular}
\end{adjustbox}
\caption{SNLI-VE stability under reordering and repetition. Accept flip counts test questions with any acceptance disagreement across the original and up to three successful reorderings, using a frozen nominal 5\% risk threshold. Answer change compares the top answer in two unchanged API calls on a separate repeat subset (successful pairs only). $n$ counts questions, not API calls. SFT is averaged over three seeds; dashes mean not measured. The rates use different events and samples and cannot be subtracted to isolate order effects.}
\label{tab:v3apistability}
\end{table}

%% file: generated/v3/tab_v3_quality.tex
\begin{table*}[t]
\centering
\small
\setlength{\tabcolsep}{4pt}
\begin{adjustbox}{max width=\linewidth}
\begin{tabular}{lccccccccc}
\toprule
 & \multicolumn{4}{c}{Accuracy} & \multicolumn{5}{c}{Macro over the four benchmarks} \\
\cmidrule(lr){2-5}\cmidrule(lr){6-10}
System & GQA & SNLI-VE & TextVQA$^\dagger$ & TallyQA$^\dagger$ & Acc$\uparrow$ & NLL$\downarrow$ & Brier$\downarrow$ & ECE$\downarrow$ & AURC$\downarrow$ \\
\midrule
Original 4B (M0) & 0.880 & 0.629 & 0.974 & 0.853 & 0.834 & 1.009 & 0.294 & 0.134 & 0.079 \\
\quad + TS (M2) & 0.880 & 0.629 & 0.974 & 0.853 & 0.834 & 0.416 & 0.230 & 0.029 & 0.080 \\
Answer SFT (M1) & 0.916 & 0.808 & 0.975 & 0.841 & 0.885 & 0.331 & 0.172 & 0.047 & 0.030 \\
\quad + TS (M3) & 0.916 & 0.808 & 0.975 & 0.841 & 0.885 & 0.309 & 0.165 & 0.025 & 0.030 \\
\midrule
Original 8B & 0.885 & 0.695 & 0.979 & 0.853 & 0.853 & 0.634 & 0.243 & 0.101 & 0.062 \\
\quad + TS & 0.885 & 0.695 & 0.979 & 0.853 & 0.853 & 0.376 & 0.208 & 0.022 & 0.062 \\
Answer SFT 8B & 0.919 & 0.818 & 0.979 & 0.855 & 0.893 & 0.298 & 0.158 & 0.037 & 0.027 \\
\quad + TS & 0.919 & 0.818 & 0.979 & 0.855 & 0.893 & 0.285 & 0.154 & 0.021 & 0.027 \\
\bottomrule
\end{tabular}
\end{adjustbox}
\caption{Answer SFT raises test accuracy on task families included in our SFT, and TS improves calibration for every model without changing accuracy. Locked test images; TS = temperature scaling fitted per checkpoint and benchmark on the calibration images; it leaves accuracy unchanged. Post-trained rows are means over three seeds. $^\dagger$Task excluded from our SFT; all accuracies are on held-out test images. TallyQA is the full-test build with images shared with GQA training removed (\Cref{sec:setup}).}
\label{tab:v3quality}
\end{table*}

%% file: generated/v3/tab_v3_ranking.tex
\begin{table*}[t]
\centering
\small
\setlength{\tabcolsep}{4pt}
\begin{adjustbox}{max width=\linewidth}
\begin{tabular}{lccccccc}
\toprule
System & AURC & E-AURC & AUROC & GQA & SNLI-VE & TextVQA$^\dagger$ & TallyQA$^\dagger$ \\
\midrule
Original 4B (M2) & 0.080 & 0.055 & 0.834 & 0.833 & 0.646 & 0.985 & 0.871 \\
Answer SFT (M3) & 0.030 & 0.021 & 0.890 & 0.892 & 0.792 & 0.984 & 0.892 \\
\midrule
Full-vocab.\ CE cont. & 0.031 & 0.022 & 0.887 & 0.885 & 0.792 & 0.984 & 0.888 \\
Cand.\ CE cont. & 0.031 & 0.022 & 0.887 & 0.884 & 0.792 & 0.984 & 0.889 \\
Cand.\ Brier cont. & 0.030 & 0.021 & 0.890 & 0.890 & 0.796 & 0.984 & 0.891 \\
Cand.\ CE + shuffle & 0.031 & 0.022 & 0.885 & 0.884 & 0.785 & 0.985 & 0.886 \\
\midrule
Original 8B & 0.062 & 0.045 & 0.851 & 0.854 & 0.669 & 0.988 & 0.892 \\
Answer SFT 8B & 0.027 & 0.019 & 0.896 & 0.906 & 0.794 & 0.988 & 0.895 \\
\bottomrule
\end{tabular}
\end{adjustbox}
\caption{Answer SFT ranks errors better even after accuracy is discounted. E-AURC subtracts the AURC of a perfect ranking at the same accuracy; AUROC is the probability that a correct answer receives higher confidence than a wrong one (half credit for ties), conditioning on correctness rather than weighting by the overall error rate. Test images, task-local TS, macro over the four benchmarks; per-benchmark columns are AUROC.}
\label{tab:v3ranking}
\end{table*}

%% file: generated/v3/tab_v3_transfer.tex
\begin{table}[t]
\centering
\footnotesize
\setlength{\tabcolsep}{4pt}
\begin{adjustbox}{max width=\linewidth}
\begin{tabular}{lcccc}
\toprule
System & NLL local & NLL transfer & ECE local & ECE transfer \\
\midrule
Original 4B & 0.416 & 0.487 & 0.029 & 0.097 \\
Answer SFT & 0.309 & 0.319 & 0.025 & 0.037 \\
\bottomrule
\end{tabular}
\end{adjustbox}
\caption{A temperature transferred from the trained tasks calibrates worse than a task-local one. Task-local TS (fitted on each benchmark's own calibration images) against transfer TS (one temperature fitted on GQA and SNLI-VE calibration images, frozen onto all four). Macro, test images.}
\label{tab:v3transfer}
\end{table}

%% file: generated/v3/tab_v3_serving.tex
\begin{table*}[t]
\centering
\small
\setlength{\tabcolsep}{4pt}
\begin{adjustbox}{max width=\linewidth}
\begin{tabular}{lcccccccc}
\toprule
 & \multicolumn{4}{c}{$N{=}8$} & \multicolumn{4}{c}{$N{=}32$} \\
\cmidrule(lr){2-5}\cmidrule(lr){6-9}
Path & ms/q & p95 ms & KV MiB & Groups & ms/q & p95 ms & KV MiB & Groups \\
\midrule
S0 HF, full sequences, batched & 20.8 & 212 & 305 & 18 & 20.5 & 836 & 1218 & 4 \\
S1 HF, shared prefix, batched & 11.4 & 111 & 336 & 17 & 5.6 & 202 & 1249 & 4 \\
S2 vLLM, no prefix caching & 18.6 & 184 & 315 & 18 & 19.0 & 789 & 1257 & 4 \\
S3 vLLM, prefix caching (cold) & 11.4 & 112 & 99 & 71 & 7.0 & 276 & 304 & 25 \\
\quad prefix already cached & 9.2 & 93 & 99 & 71 & 6.9 & 267 & 304 & 25 \\
\bottomrule
\end{tabular}
\end{adjustbox}
\caption{HF prefix sharing is 3.7$\times$ faster at $N{=}32$; engine caching uses 24.3\% of S1's KV memory. One image group at a time, same GPU, weights (merged answer SFT, bf16) and natural GQA questions. ms/q is group time divided by $N$. KV is the logical key--value footprint of one group from its real token counts (144\,KiB per token; vLLM rounded to 16-token blocks); Groups is how many such groups fit in an 8\,GiB KV budget.}
\label{tab:v3serving}
\end{table*}

%% file: generated/v3/tab_v3_load.tex
\begin{table}[t]
\centering
\footnotesize
\setlength{\tabcolsep}{4pt}
\begin{adjustbox}{max width=\linewidth}
\begin{tabular}{lcccccc}
\toprule
 & \multicolumn{3}{c}{$N{=}8$} & \multicolumn{3}{c}{$N{=}32$} \\
\cmidrule(lr){2-4}\cmidrule(lr){5-7}
Path & Max g/s & p50 & p95 & Max g/s & p50 & p95 \\
\midrule
S1 HF, shared prefix & 10.9 & 196 & 648 & 5.6 & 209 & 497 \\
S2 vLLM, no caching & 9.1 & 459 & 1114 & 2.3 & 10344 & 23116 \\
S3 vLLM, caching & 16.2 & 93 & 171 & 4.9 & 304 & 671 \\
\bottomrule
\end{tabular}
\end{adjustbox}
\caption{Under load, vLLM with prefix caching sustains the most groups at $N{=}8$ and our shared path at $N{=}32$. Image groups arrive on one fixed Poisson schedule (200 per rate) and latency runs from arrival to the group's last answer, queueing included. Max g/s is the highest achieved group throughput over the offered rates; p50/p95 (ms) are at 8 groups/s for $N{=}8$ and 3 groups/s for $N{=}32$. S1 serves one group at a time; vLLM batches across groups in flight.}
\label{tab:v3load}
\end{table}

%% file: generated/v3/tab_v3_selective_ci.tex
\begin{table}[t]
\centering
\footnotesize
\setlength{\tabcolsep}{4pt}
\begin{adjustbox}{max width=\linewidth}
\begin{tabular}{lllcc}
\toprule
Benchmark & Target & System & Coverage & Realized risk [95\% CI] \\
\midrule
GQA & 5\% & M2 & 0.758 & 0.047 [0.041, 0.054] \\
 &  & M3 & 0.888 & 0.044 [0.039, 0.050] \\
 & 10\% & M2 & 0.936 & 0.093 [0.085, 0.101] \\
 &  & M3 & 0.999 & 0.083 [0.076, 0.090] \\
\midrule
SNLI-VE & 5\% & M2 & 0.023 & 0.125 [0.029, 0.259] \\
 &  & M3 & 0.346 & 0.038 [0.025, 0.054] \\
 & 10\% & M2 & 0.067 & 0.076 [0.029, 0.138] \\
 &  & M3 & 0.626 & 0.079 [0.063, 0.096] \\
\midrule
TextVQA$^\dagger$ & 5\% & M2 & 1.000 & 0.026 [0.019, 0.034] \\
 &  & M3 & 1.000 & 0.025 [0.017, 0.031] \\
 & 10\% & M2 & 1.000 & 0.026 [0.019, 0.034] \\
 &  & M3 & 1.000 & 0.025 [0.017, 0.031] \\
\midrule
TallyQA$^\dagger$ & 5\% & M2 & 0.748 & 0.047 [0.039, 0.055] \\
 &  & M3 & 0.764 & 0.053 [0.046, 0.061] \\
 & 10\% & M2 & 0.898 & 0.099 [0.089, 0.111] \\
 &  & M3 & 0.887 & 0.103 [0.093, 0.114] \\
\bottomrule
\end{tabular}
\end{adjustbox}
\caption{Frozen-threshold deployment with image-cluster bootstrap intervals for realized risk under the ranked-prefix cutoff. Intervals describe test-image uncertainty; overlap with a target does not establish a risk guarantee.}
\label{tab:v3selci}
\end{table}

%% file: generated/v3/tab_v3_tally.tex
\begin{table*}[t]
\centering
\small
\setlength{\tabcolsep}{4pt}
\begin{adjustbox}{max width=\linewidth}
\begin{tabular}{lccccccc}
\toprule
Build & $n$ test & Acc M0 & Acc M1 & Risk@5\% M2 & Risk@5\% M3 & Risk@10\% M2 & Risk@10\% M3 \\
\midrule
Short, count-balanced & 338 & 0.340 & 0.345 & 0.185 & 0.108 & 0.203 & 0.235 \\
Full test, natural & 3,681 & 0.853 & 0.841 & 0.047 & 0.053 & 0.099 & 0.103 \\
\bottomrule
\end{tabular}
\end{adjustbox}
\caption{TallyQA results depend on the evaluation build. The short, count-balanced subset has 160 calibration questions; the larger build follows the natural count distribution. Both sample size and distribution change, so this comparison does not isolate their effects.}
\label{tab:v3tally}
\end{table*}

%% file: generated/v3/tab_v3_contrast_full.tex
\begin{table*}[t]
\centering
\footnotesize
\setlength{\tabcolsep}{4pt}
\begin{adjustbox}{max width=\linewidth}
\begin{tabular}{lcccc}
\toprule
Contrast (macro) & $\Delta$Acc & $\Delta$NLL & $\Delta$ECE & $\Delta$AURC \\
\midrule
Answer SFT $-$ original (raw) & +0.051 {\scriptsize [0.044, 0.059]} & $-$0.678 {\scriptsize [$-$0.737, $-$0.626]} & $-$0.088 {\scriptsize [$-$0.094, $-$0.080]} & $-$0.049 {\scriptsize [$-$0.056, $-$0.042]} \\
TS on the original & +0.000 {\scriptsize [0.000, 0.000]} & $-$0.593 {\scriptsize [$-$0.645, $-$0.546]} & $-$0.105 {\scriptsize [$-$0.111, $-$0.096]} & +0.001 {\scriptsize [0.001, 0.002]} \\
TS on answer SFT & +0.000 {\scriptsize [0.000, 0.000]} & $-$0.022 {\scriptsize [$-$0.027, $-$0.017]} & $-$0.022 {\scriptsize [$-$0.026, $-$0.014]} & $-$0.000 {\scriptsize [$-$0.000, $-$0.000]} \\
Answer SFT $-$ original (TS) & +0.051 {\scriptsize [0.044, 0.059]} & $-$0.107 {\scriptsize [$-$0.117, $-$0.096]} & $-$0.004 {\scriptsize [$-$0.011, 0.004]} & $-$0.050 {\scriptsize [$-$0.057, $-$0.044]} \\
\midrule
Typed head $-$ LM head (raw) & $-$0.005 {\scriptsize [$-$0.007, $-$0.002]} & +0.085 {\scriptsize [0.074, 0.098]} & +0.027 {\scriptsize [0.024, 0.030]} & +0.003 {\scriptsize [0.002, 0.005]} \\
Typed head $-$ LM head (TS) & $-$0.005 {\scriptsize [$-$0.007, $-$0.002]} & +0.029 {\scriptsize [0.024, 0.033]} & +0.003 {\scriptsize [$-$0.001, 0.006]} & +0.003 {\scriptsize [0.002, 0.005]} \\
\midrule
Full-vocab.\ CE cont.\ $-$ M3 & $-$0.002 {\scriptsize [$-$0.004, 0.000]} & +0.011 {\scriptsize [0.008, 0.013]} & +0.006 {\scriptsize [0.001, 0.008]} & +0.001 {\scriptsize [0.001, 0.002]} \\
Cand.\ CE $-$ full-vocab.\ CE (TS) & +0.000 {\scriptsize [$-$0.001, 0.001]} & +0.000 {\scriptsize [0.000, 0.001]} & $-$0.000 {\scriptsize [$-$0.002, 0.001]} & $-$0.000 {\scriptsize [$-$0.000, 0.000]} \\
Cand.\ Brier $-$ full-vocab.\ CE (TS) & +0.000 {\scriptsize [$-$0.001, 0.001]} & $-$0.005 {\scriptsize [$-$0.006, $-$0.004]} & $-$0.002 {\scriptsize [$-$0.004, 0.001]} & $-$0.001 {\scriptsize [$-$0.001, $-$0.000]} \\
Cand.\ CE+shuffle $-$ full-vocab.\ CE (TS) & +0.002 {\scriptsize [$-$0.000, 0.003]} & $-$0.002 {\scriptsize [$-$0.004, $-$0.001]} & $-$0.000 {\scriptsize [$-$0.003, 0.003]} & +0.000 {\scriptsize [$-$0.000, 0.000]} \\
Cand.\ Brier $-$ cand.\ CE (TS) & +0.000 {\scriptsize [$-$0.001, 0.001]} & $-$0.005 {\scriptsize [$-$0.006, $-$0.004]} & $-$0.002 {\scriptsize [$-$0.004, 0.001]} & $-$0.001 {\scriptsize [$-$0.001, $-$0.000]} \\
Cand.\ CE+shuffle $-$ cand.\ CE (TS) & +0.002 {\scriptsize [0.000, 0.003]} & $-$0.003 {\scriptsize [$-$0.004, $-$0.001]} & $-$0.000 {\scriptsize [$-$0.002, 0.003]} & +0.000 {\scriptsize [$-$0.000, 0.001]} \\
\bottomrule
\end{tabular}
\end{adjustbox}
\caption{All paired contrasts, macro over the four benchmarks, with 95\% image-cluster bootstrap intervals. Typed head and LM head are trained on the same varied-$K$ corpus and budget.}
\label{tab:v3contrastfull}
\end{table*}

%% file: generated/v3/tab_v3_byk.tex
\begin{table}[t]
\centering
\footnotesize
\setlength{\tabcolsep}{4pt}
\begin{adjustbox}{max width=\linewidth}
\begin{tabular}{lccccccc}
\toprule
$K$ & $n$ & M0 Acc & M0 gap raw & M0 gap TS & M1 Acc & M1 gap raw & M1 gap TS \\
\midrule
2 & 2946 & 0.866 & +0.099 & +0.025 & 0.913 & +0.051 & +0.006 \\
3 & 523 & 0.935 & +0.047 & $-$0.024 & 0.957 & +0.020 & $-$0.011 \\
4 & 487 & 0.893 & +0.088 & +0.005 & 0.932 & +0.037 & $-$0.001 \\
5 & 505 & 0.887 & +0.088 & $-$0.003 & 0.910 & +0.059 & +0.019 \\
6 & 498 & 0.896 & +0.084 & $-$0.019 & 0.916 & +0.048 & +0.003 \\
8 & 488 & 0.869 & +0.097 & $-$0.024 & 0.886 & +0.068 & +0.010 \\
\bottomrule
\end{tabular}
\end{adjustbox}
\caption{Both models are overconfident at each tested option count; TS reduces the gaps without a dependence on $K$. GQA test questions by option count $K$. Gap = mean confidence $-$ accuracy (positive = overconfident). One temperature per checkpoint and benchmark leaves no systematic dependence on $K$.}
\label{tab:v3byk}
\end{table}

%% file: generated/v3/tab_v3_ensemble.tex
\begin{table}[t]
\centering
\footnotesize
\setlength{\tabcolsep}{4pt}
\begin{adjustbox}{max width=\linewidth}
\begin{tabular}{llcccccc}
\toprule
System & Variant & Acc & NLL & ECE & AURC & Cov@5\% & Risk@5\% \\
\midrule
Original 4B & one order & 0.706 & 0.698 & 0.050 & 0.185 & 0.485 & 0.096 \\
 & 4-order average & 0.710 & 0.665 & 0.038 & 0.173 & 0.494 & 0.079 \\
Answer SFT & one order & 0.761 & 0.591 & 0.043 & 0.131 & 0.597 & 0.054 \\
 & 4-order average & 0.756 & 0.580 & 0.045 & 0.130 & 0.613 & 0.068 \\
\bottomrule
\end{tabular}
\end{adjustbox}
\caption{Averaging over four option orders buys little for four times the inference. Averaging the semantic distribution over the original order and three reorderings (four forward passes) against the single order, both with frozen TS; thresholds for the average are refitted on its own calibration distributions. Macro over GQA, SNLI-VE, TextVQA and the short TallyQA subset.}
\label{tab:v3ens}
\end{table}

%% file: generated/v3/tab_v3_parity.tex
\begin{table}[t]
\centering
\footnotesize
\setlength{\tabcolsep}{4pt}
\begin{adjustbox}{max width=\linewidth}
\begin{tabular}{llccccc}
\toprule
Weights & APC & Flips & TV p50 & TV p95 & Acc HF & Acc vLLM \\
\midrule
Original 4B & off & 18/3,319 & 3.9e-06 & 0.046 & 0.820 & 0.822 \\
 & on & 18/3,319 & 4.2e-06 & 0.052 & 0.820 & 0.821 \\
Answer SFT (merged) & off & 11/3,319 & 3.1e-05 & 0.027 & 0.867 & 0.868 \\
 & on & 15/3,319 & 3.1e-05 & 0.029 & 0.867 & 0.867 \\
\bottomrule
\end{tabular}
\end{adjustbox}
\caption{vLLM and HF Transformers differ on fewer than 0.6\% of argmax decisions overall. vLLM (0.30) against HF on the same prompts and weights, calibration images of all four benchmarks. vLLM returns the candidate logits themselves (processed logits under an allowed-token mask); the candidate distribution is formed identically on both sides. APC = automatic prefix caching.}
\label{tab:v3parity}
\end{table}

%% file: generated/v3/tab_v3_s4.tex
\begin{table}[t]
\centering
\footnotesize
\setlength{\tabcolsep}{4pt}
\begin{adjustbox}{max width=\linewidth}
\begin{tabular}{llcccc}
\toprule
Weights & Interface & Acc & NLL TS & ECE TS & AURC \\
\midrule
Original 4B & \texttt{Answer:} readout & 0.706 & 0.698 & 0.050 & 0.185 \\
 & \texttt{/v1/systemone} & 0.717 & 0.675 & 0.041 & 0.173 \\
\midrule
Answer SFT (seed 0) & \texttt{Answer:} readout & 0.758 & 0.597 & 0.043 & 0.137 \\
 & \texttt{/v1/systemone} & 0.761 & 0.594 & 0.044 & 0.135 \\
\bottomrule
\end{tabular}
\end{adjustbox}
\caption{The answer-SFT adapter achieves macro accuracy 0.761 through the native prompt, against 0.758 with our readout. Both prompts use identical weights; vLLM's native endpoint builds its prompt from the model's chat template. Macro over GQA, SNLI-VE, TextVQA and the short TallyQA subset (the endpoint was not run on the full-test build).}
\label{tab:v3s4}
\end{table}

%% file: txt_repro_cvpr.tex
\paragraph{Model and training.} The main system uses Qwen3-VL-4B-Instruct in
\texttt{bfloat16}, a frozen vision tower, and LoRA ($r{=}16$,
$\alpha{=}32$, dropout $0.05$) on the language tower's attention and MLP
projections. Answer SFT trains no additional head; the typed-head control adds Choice and
Claim heads in \texttt{float32}. AdamW uses a cosine schedule
with 100 warm-up steps, learning rate $10^{-4}$ for LoRA and $10^{-3}$ for the
heads, gradient clipping at $1.0$, and batch size 8 with gradient
checkpointing. Each run uses \VDMsteps{} updates on one GPU. The 8B comparison
changes only backbone size.

\paragraph{Image resolution.} Training, evaluation, and timing in both HF
and vLLM use the image processor's default resolution: a $500\times500$ GQA
image becomes 256 visual tokens. The configured cap of 196 visual tokens
($448\times448$) uses an attribute ignored by the Transformers release;
all reported paths therefore use the same default processing.

\paragraph{Probability-targeted continuations.} Each released seed is
continued for 1{,}000 updates at learning rate $2\times10^{-5}$ with 50 warm-up
steps, the same data (the fixed-$K$ GQA build and SNLI-VE), LoRA parameters,
batch and schedule shape. The four arms differ only in the objective; no arm
was tuned. Candidate CE and Brier are computed on the LM-head logits of the
option letters, normalized over each item's own options; the shuffled arm
draws a fresh option order for every example and moves the label with it.

\paragraph{Calibration and thresholds.} TS fits one temperature per checkpoint
and benchmark by golden-section search on calibration NLL. The reported
cutoff selects the largest confidence-ranked calibration prefix whose error
is at most the target; stable sorting breaks ties by item order. If no prefix
qualifies, the system accepts nothing. Otherwise it accepts all examples
at or above the last prefix confidence, including ties beyond the selected
prefix.

\paragraph{Hardware and software.} One NVIDIA RTX 5090 (32\,GB) per run,
PyTorch 2.14 with CUDA 13.0, Transformers 5.17, vLLM 0.30 for the engine
paths and a development build of vLLM with the decision endpoint for S4.
Training a single variant takes about 45 minutes; a continuation about 15.

\paragraph{Decision APIs.} Both APIs were queried in November 2026 with
pinned model versions. Each request holds the state ``Answer from the attached
image.'' followed by the image, and one choice question per item whose
instructions are the question text and whose options are the displayed
options, in order, without descriptions; SNLI-VE uses its three claim options.
Option reorderings are sent as separate requests, one per reordering. Failed
requests were retried up to five times; the raw responses are kept with the
predictions. The repeat check re-sent the first 1{,}500 original GQA and
SNLI-VE questions and all short-TallyQA questions unchanged.

\paragraph{Splits.} The loader trains only on source-training items marked
\texttt{train}. Evaluation uses GQA source validation and SNLI-VE source test,
removing SNLI-VE images shared with our training set. Variants retain the
parent split. Within evaluation pools, a fixed parent-image hash assigns 30\%
to calibration and the rest to test, identically for all systems. This is
separate from the SFT training split; temperatures and thresholds use only
calibration images.

\paragraph{Reproduction.} Every run writes its configuration, seed and raw
per-example predictions; a manifest records the hash of every checkpoint and
prediction file each table reads. All tables and every number quoted in the
text are generated from those files by script; none are transcribed by hand.

%% file: generated/v3/tab_v3_api.tex
\begin{table*}[t]
\centering
\small
\setlength{\tabcolsep}{4pt}
\begin{adjustbox}{max width=\linewidth}
\begin{tabular}{lccccccccccc}
\toprule
& \multicolumn{5}{c}{Test accuracy} & \multicolumn{2}{c}{NLL} & \multicolumn{2}{c}{ECE} & & \\
\cmidrule(lr){2-6}\cmidrule(lr){7-8}\cmidrule(lr){9-10}
System & GQA & SNLI-VE & TextVQA$^\dagger$ & TallyQA$^\dagger$ & Macro & raw & TS & raw & TS & AUROC & Failed \\
\midrule
Visual Jev (untrained) & 0.880 & 0.629 & 0.974 & 0.853 & 0.834 & 1.009 & 0.416 & 0.134 & 0.029 & 0.834 & -- \\
Visual Jev (trained) & 0.916 & 0.808 & 0.975 & 0.841 & 0.885 & 0.331 & 0.309 & 0.047 & 0.025 & 0.890 & -- \\
Jev API & 0.885 & 0.713 & 0.982 & 0.851 & 0.858 & 0.385 & 0.361 & 0.048 & 0.013 & 0.870 & 1.9\% \\
OpenAI Decisions & 0.834 & 0.732 & 0.955 & 0.818 & 0.835 & 0.486 & 0.440 & 0.060 & 0.020 & 0.833 & 0.3\% \\
\bottomrule
\end{tabular}
\end{adjustbox}
\caption{Held-out test performance on identical questions, option orders and splits. Trained/untrained = M1/M0, with/without task SFT of the pretrained 4B; $^\dagger$marks tasks excluded from SFT. API training data are unknown. NLL, ECE and AUROC are macro averages; AUROC after TS. Failed: question share assigned uniform probabilities after request failure.}
\label{tab:v3api}
\end{table*}

%% file: generated/v3/tab_v3_d2.tex
\begin{table}[t]
\centering
\footnotesize
\setlength{\tabcolsep}{4pt}
\begin{adjustbox}{max width=\linewidth}
\begin{tabular}{lcccccc}
\toprule
Continuation & \multicolumn{2}{c}{NLL} & ECE & AURC & \multicolumn{2}{c}{SNLI order} \\
\cmidrule(lr){2-3}\cmidrule(lr){6-7}
objective & raw & TS & TS & & same & flip \\
\midrule
None (M1/M3) & 0.331 & 0.309 & 0.025 & 0.030 & 0.926 & 0.170 \\
Full-vocab.\ CE & 0.441 & 0.320 & 0.031 & 0.031 & 0.922 & 0.167 \\
Cand.\ CE & 0.439 & 0.320 & 0.031 & 0.031 & 0.920 & 0.172 \\
Cand.\ Brier & 0.357 & 0.315 & 0.029 & 0.030 & 0.918 & 0.168 \\
Cand.\ CE + shuffle & 0.423 & 0.317 & 0.031 & 0.031 & 0.938 & 0.163 \\
\bottomrule
\end{tabular}
\end{adjustbox}
\caption{None of the tested continuations lowers macro NLL or ECE after TS relative to answer SFT plus TS. Each released answer-SFT seed is continued for 1{,}000 further updates on the same data, LoRA, batch and schedule, differing only in the objective; no per-arm tuning. Macro over the four benchmarks; SNLI columns are the option-order statistics of \Cref{tab:v3order}.}
\label{tab:v3d2}
\end{table}

%% file: txt_limitations_cvpr.tex
\paragraph{Empirical thresholds and dataset construction.}
Thresholds use labeled calibration images and have no finite-sample risk
guarantee. The cutoff selects ranked prefixes but admits complete tie groups
at deployment, so ties can raise calibration risk above the nominal target.
Reported intervals capture test-image sampling,
not uncertainty from refitting calibration or training new seeds. The short
and full TallyQA builds differ in both size and count distribution, so they
do not isolate the effect of calibration-set size. Probabilities are
conditional on the supplied candidates and do not detect a missing correct
answer.

\paragraph{Stability and deployment scope.}
Three fixed reorderings probe a limited part of the permutation space. The
counting order study uses the short subset. Acceptance inconsistency does not
identify the resulting excess risk or review cost; those require directional
error analysis. API repeat checks use a subset and two calls per request, and
do not separate model randomness, routing, caching, or probability rounding.
Failures represented as uniform distributions are an evaluation convention,
not actual API answers. Hosted model training and execution are unknown.

\paragraph{Training and generalization.}
Probability-targeted continuations use one untuned 1,000-update budget and
three seeds. Controlled training comparisons use one backbone
family at two scales, a frozen vision tower, English questions, and at most
\VDMKmax{} candidates. Gains concentrate on the trained families; TextVQA is
near ceiling and 4B counting accuracy declines. The hosted APIs receive no
task-specific fine-tuning in this study. The comparison evaluates the served
models on the same held-out questions; pretraining exposure and provider-side
adaptation are undisclosed.

\paragraph{Serving and visual evidence.}
Serving uses one RTX 5090 on a shared machine, 200 arrivals per load condition,
and $N\in\{8,32\}$. The native endpoint uses one seed and short TallyQA.
With grey images, GQA accuracy is \VDMgqaBlind{} against \VDMgqaChance{}
chance, and SNLI-VE \VDMsnliBlind{} against \VDMsnliChance{}.
Candidate priors therefore contribute to benchmark accuracy.